\documentclass[10pt,journal,compsoc,twoside]{IEEEtran}
\ifCLASSOPTIONcompsoc
  \usepackage[nocompress]{cite}
\else
  \usepackage{cite}
\fi

\usepackage{caption}

\ifCLASSINFOpdf
  \usepackage[pdftex]{graphicx}
\else
\fi

\usepackage{amsmath}

\usepackage{amssymb}
\usepackage[table,xcdraw]{xcolor}

\ifCLASSOPTIONcompsoc
  \usepackage[caption=false,font=footnotesize,labelfont=sf,textfont=sf]{subfig}
\else
  \usepackage[caption=false,font=footnotesize]{subfig}
\fi

\usepackage{multirow}   

\usepackage{booktabs}

\begin{document}
\title{Complex-valued Iris Recognition Network}

%
%
%
%

\author{Kien~Nguyen,~\IEEEmembership{Member,~IEEE,}
        Clinton~Fookes,~\IEEEmembership{Senior Member,~IEEE,}
        Sridha~Sridharan,~\IEEEmembership{Life Senior Member,~IEEE,}        
        and~Arun~Ross,~\IEEEmembership{Senior Member,~IEEE,}                
\IEEEcompsocitemizethanks{\IEEEcompsocthanksitem K. Nguyen, C. Fookes and S. Sridharan are with Image and Video Research Laboratory, SAIVT, School of Electrical Engineering and Computer Science, Queensland University of Technology, Brisbane, QLD, 4000, Australia.\protect\\
E-mail: {k.nguyenthanh,c.fookes,s.sridharan}@qut.edu.au}

\IEEEcompsocitemizethanks{\IEEEcompsocthanksitem Arun Ross is with the Department of Computer Science and Engi- neering, Michigan State University, East Lansing, MI 48824 USA. \protect\\
Email: rossarun@cse.msu.edu}

}

%
%

\markboth{IEEE Transactions}%
{Nguyen \MakeLowercase{\emph{et al.}}: Complex-valued Iris Recognition Network}
%



\IEEEtitleabstractindextext{%
\begin{abstract}

In this work, we design a fully complex-valued neural network for the task of iris recognition. Unlike the problem of general object recognition, where real-valued neural networks can be used to extract pertinent features, iris recognition depends on the extraction of both phase and magnitude information from the input iris texture in order to better represent its biometric content. This necessitates the extraction and processing of phase information that cannot be effectively handled by a real-valued neural network. In this regard, we design a fully complex-valued neural network that can better capture the multi-scale, multi-resolution, and multi-orientation phase and amplitude features of the iris texture. We show a strong correspondence of the proposed complex-valued iris recognition network with Gabor wavelets that are used to generate the classical IrisCode; however, the proposed method enables a new capability of automatic complex-valued feature learning that is tailored for iris recognition. We conduct experiments on three benchmark datasets - ND-CrossSensor-2013, CASIA-Iris-Thousand and UBIRIS.v2 - and show the benefit of the proposed network for the task of iris recognition. We exploit visualization schemes to convey how the complex-valued network, when compared to standard real-valued networks, extracts fundamentally different features from the iris texture. 

\end{abstract}

}

\maketitle

\IEEEdisplaynontitleabstractindextext

%
\IEEEpeerreviewmaketitle

\IEEEraisesectionheading{\section{Introduction}\label{sec:introduction}}
\IEEEPARstart{T}{he} human iris is a powerful biometric pattern that has the potential to deliver high recognition accuracy at low false match rates. This is due to (i) the complex textural pattern of the iris  that is believed to be unique to each eye, and (ii) the limited genetic penetrance of the iris texture~\cite{IrisStructure,DaugmanInformationTheory,Ross2010iris}. The success of iris recognition - besides its attractive physical characteristics - is rooted in the development of efficient feature descriptors, especially the IrisCode introduced in Daugman's pioneering work \cite{DaugmanFirstPaper,Daugman07,DaugmanInformationTheory} and many other descriptors that have subsequently evolved ~\cite{IrisCodeAnalysis,irisPhaseDFT,irisOrdinalMeasures,IrisZM,LongRangeIris,IrisCompressiveSensing,OptimalIrisCode}.

The last few years have seen a transition in the iris recognition community to deep neural networks to take advantage of their \emph{automatic feature learning} capability \cite{DeepIris,IrisFCN,DRFNet,ConstrainedIrisNet}. This provides us with an alternative approach in feature design by automatically learning and discovering feature representations directly from data, eliminating some of the pitfalls in developing handcrafted features \cite{RepLearning}. Despite the promise and a number of initial efforts in this direction, deep neural networks have not been exactly revolutionary in iris recognition. This could be because existing deep iris recognition networks in the literature are directly derived from general deep learning theory for natural images. These approaches do not take into account some of the unique properties of the iris texture. Compared to the statistics of natural images as defined in \cite{NaturalScenes}, the iris texture itself is stochastic \cite{IrisComplexity} without consistent shapes, edges, or stromal morphology. This makes it significantly different from non-stochastic and structured patterns occurring in object-based natural images. The \emph{intrinsic differences between iris texture images and object-based natural images} requires automatic feature learning to be tailored with domain-specific knowledge in order for it to reach its full potential in the iris recognition setting. 

\begin{figure*}
\centering
\includegraphics[width=2\columnwidth]{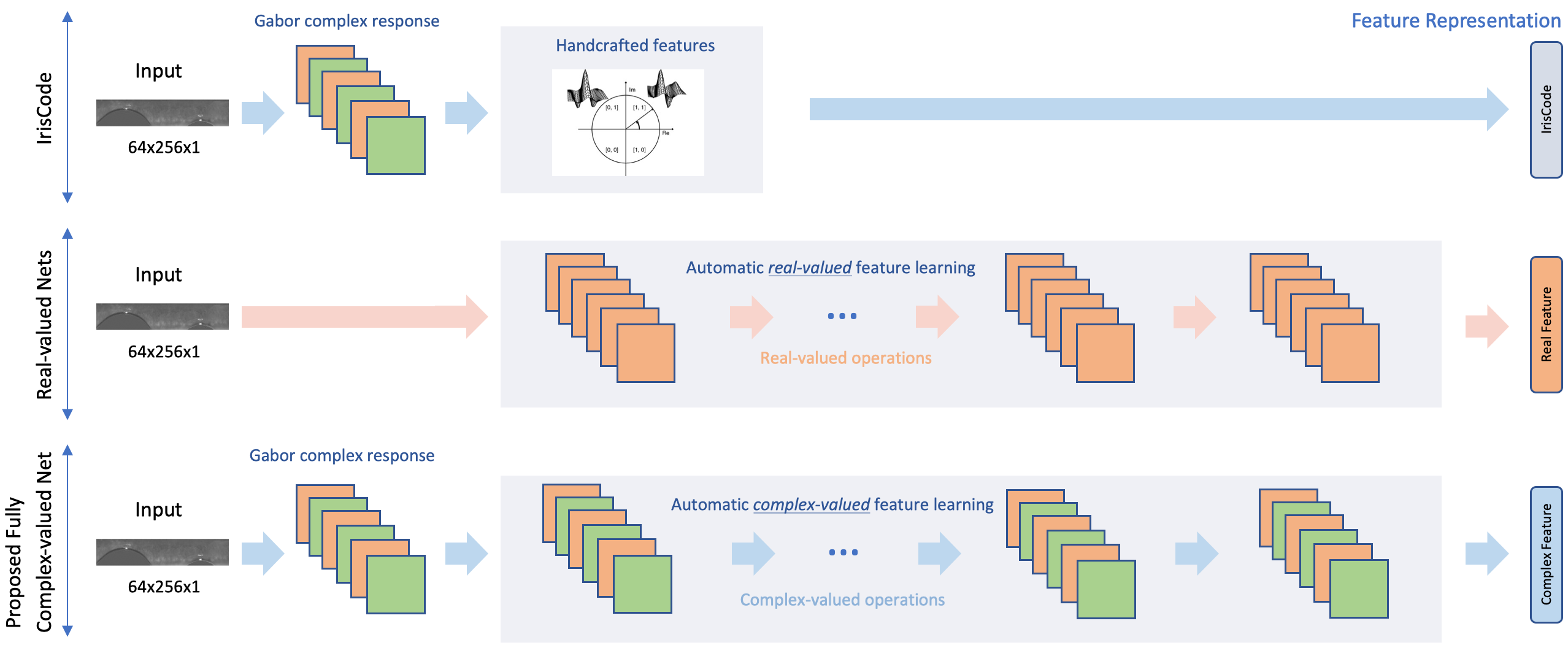}
\caption{The classic IrisCode depends on the complex-valued Gabor response followed by phase quantization, which is not learned from the data. Existing deep iris recognition networks perform automatic feature learning in the real-valued domain, which cannot adequately capture and retain phase information along the pipeline. This paper proposes to shift processing of the iris texture to a fully complex-valued space, which has a richer representation capacity, better leverages the unique characteristics of the iris texture, and has a strong correspondence with complex Gabor wavelets.}
\label{fig:architecturecomparison}
\end{figure*}

The contribution of this paper stems from our investigation of the classical IrisCode. Technically speaking, we can split the encoding process of the IrisCode into two steps: (i) representing the iris texture image in the complex space by applying Gabor wavelets, and (ii) quantizing the phase of the complex representation to generate the final descriptor. While the first step emphasizes the importance of the complex-valued representation, the second accentuates the importance of phase information. The importance of both complex-valued representation and phase information has been reinforced in many other handcrafted approaches. For example, Kong \emph{et al.} represented the iris texture image in the Gabor complex space and employed precise phase to encode the iris \cite{IrisCodeAnalysis}. Monro \emph{et al.} \cite{irisPhaseDCT} and Miyazawa \emph{et al.} \cite{irisPhaseDFT} represented the iris texture image in the Cosine and Fourier spaces and then used the phase to encode the iris.

All existing deep networks proposed for iris recognition in the literature have not been able to incorporate this domain knowledge. First, they all operate in the real-valued space with real-valued operations and feature maps. Compared to a complex-valued representation, existing real-valued iris recognition networks may be limited in the space in which they can learn feature representations. The fact that they completely ignore the \emph{complex-valued nature} of the iris feature representation may discard many of the benefits of the complex-valued representation. Second, all existing deep iris recognition networks have no mechanism to explicitly retain the phase along the network pipeline. Explicitly modeling and processing phase information in the automatic feature learning process is expected to better deal with the unique properties of the iris texture.

Motivated by these insights, we propose to shift from standard automatic real-valued feature learning as in current deep iris networks to \emph{automatic complex-valued feature learning}. On the one hand, the shift from a real-valued space to a complex-valued space will enrich the search space for the feature learning process. On the other hand, the space shift allows us to incorporate domain knowledge of iris recognition to tailor modern deep learning to take into account the unique properties of the iris texture.

Applying complex-valued networks in the iris recognition setting is not obvious, since the iris images themselves are real numbers, which explains why all existing deep iris networks have been designed and operated only in the real domain. We re-purpose the complex response of Gabor wavelets as the complex input to be further processed by a fully complex-valued network. This enables us to retain the complex-valued nature of the feature representation through the entire pipeline. When the complex-valued nature of the feature representation is retained, it is trivial to recover the phase information from the real and imaginary components. Compared to the classical IrisCode and its phase-based derivatives, the proposed fully complex-valued iris recognition network retains the benefits of operating in the complex space, but learns a feature representation in a data-driven manner as opposed to being human-designed. This is illustrated in Fig.~\ref{fig:architecturecomparison}.

The core contributions of this work can be divided into three parts.
\begin{itemize}
\item We propose to shift processing of the iris texture to the complex-valued space in deep neural networks. To the best of our knowledge, this is the first attempt of its kind. The complex-valued concepts allow the network to better cater to the unique properties of iris features, explicitly retaining the phase information in the feature representation. However, it also requires fundamental changes to the network design and training process. 

\item We propose a novel fully complex-valued network to enable this shift. It enables, for the first time, automatic complex-valued feature learning in the iris recognition setting. Compared to standard automatic feature learning in standard real-valued networks, automatic complex-valued feature learning discovers more discriminative representations for the iris texture.

\item We provide an insight that fully complex-valued networks have a solid mathematical and theoretical foundation to better suit iris recognition than standard real-valued networks. In addition, complex-valued networks show a strong correspondence with the classic IrisCode and its phase-based derivatives. This allows us to validate the optimality of the existing handcrafted feature representations. 

\end{itemize}


The remainder of the paper is organized in 5 sections. Section~\ref{sec:Related} discusses the evolution of representation in iris recognition, from classical handcrafted approaches to modern deep learning and the need for tailored deep learning for iris recognition. Section~\ref{sec:WhyComplexNet} justifies why complex-valued networks better suit iris recognition than standard real-valued networks. Section~\ref{sec:proposed} presents our proposed complex-valued iris recognition network.
Section~\ref{sec:Experiments} describes our experimental results and the paper is concluded in Section~\ref{sec:Conclusion}.


\section{Representation evolution in iris recognition}
\label{sec:Related}
Over the last 20 years, representation of the iris texture for iris recognition has evolved from classic handcrafted features to modern representation learning using deep learning. 

\vspace{3px}
\noindent\textbf{Classic handcrafted features}\\
The richness and stochasticity of the iris texture lends itself to the application of Gabor wavelets with Hamming distance for comparing iris images \cite{DaugmanInformationTheory}. Complex-valued Gabor responses capture the rich details of the iris texture in terms of both spectral and spatially localized properties \cite{DaugmanEncyclopedia}. Both coarsely-quantized phase \cite{DaugmanFirstPaper} and precise phase \cite{IrisCodeAnalysis} of the Gabor response can create stable bit streams for a given eye, which are, nevertheless, sufficiently different across different eyes. The efficiency of phase encoding in a complex-valued space is reinforced through many other descriptors that have subsequently evolved ~\cite{IrisCodeAnalysis,irisPhaseDFT,irisPhaseDCT,IrisZM,OptimalIrisCode}. Miyazawa \emph{et al.} and Monro \emph{et al.} encoded phase components in 2D Discrete Fourier Transforms (DFTs) and 2D Discrete Cosine Transforms (DCTs) to extract a phase code for representing iris information \cite{irisPhaseDFT,irisPhaseDCT}. Similarly, Tan \emph{et al.} encoded Zernike moments-based phase features which are computed from partially overlapping regions to more effectively accommodate local pixel region variations in the normalized iris images \cite{IrisZM}. Departing from the complex-valued phase-based encoding, some notable works harness ordinal measures \cite{irisOrdinalMeasures}, compressive sensing theory \cite{IrisCompressiveSensing} and deformation field \cite{IRINA} to represent irises.


The use of handcrafted features has certain advantages. One of the biggest advantages is the use of domain knowledge gained over a period of time to focus on the biological and physical processes constituting iris recognition. For example, algorithms such as IrisCode \cite{Daugman03} are based on a deep understanding of the iris structure. Modern representation learning approaches using deep networks provide another perspective in designing iris features by automatically learning representations directly from data \cite{CNNMinutiae}.

\vspace{6px}
\noindent\textbf{Modern representation learning}\\
Deep networks have been employed to learn feature representations for iris images automatically and directly from data. There are two categories of deep networks proposed for iris recognition: classification networks and similarity networks. Classification iris networks employ deep architectures with a softmax loss to classify an input iris image into a list of known identities. Typical examples of these networks are \cite{DeepIrisNet, OTS_CNN_Iris}. The main requirement of the softmax-based networks is that the test image has to belong to one of the classes in the training set, which means the networks will have to be re-trained whenever a new class is added. In contrast, similarity iris networks employ deep architectures with a pairwise loss to learn a metric representing how similar or dissimilar two iris images are without knowing their identities. Zhao \emph{et al.} argued that classification networks may not be optimal for iris recognition since the iris texture is inherently stochastic and does not exhibit structural information or meaningful hierarchies \cite{IrisFCN}. Compared to the classification networks, similarity iris networks directly reflect what we want to achieve, \emph{i.e.,} to train the representation to correspond to iris (dis)similarity. This results in irises of the same subject having small distances and irises of different subjects having larger distances. Typical examples of these networks are \cite{DeepIris,IrisFCN,DRFNet,ConstrainedIrisNet}.

To fully exploit the capacity of deep learning in the iris recognition setting, it is required to tailor networks to suit the domain specific properties of the iris texture and for iris recognition. This paper will propose a tailored deep learning approach for iris recognition using complex-valued networks.

\section{Why complex-valued networks for iris recognition?}
\label{sec:WhyComplexNet}
Complex-valued networks originated in application domains where the input is complex-valued such as remote sensing \cite{ComplexCNN_SAR,Surreal} and MRI fingerprinting \cite{ComplexCNN_MRI}. Compared to standard real-valued networks, complex-valued networks offer three key distinctive advantages for iris recognition that can not be directly achieved by their standard real-valued counterparts. 

\vspace{3px}
\noindent\textbf{Richer representational capacity for automatic feature learning:} One key advantage of neural networks is automatic feature learning, which employs hierarchical multi-layer networks to learn a feature representation directly from data \cite{RepLearning}. The complex-valued space of complex-valued networks allows the learning algorithm to explore a richer and more versatile search space than the real-valued space of real-valued networks, potentially leading to the capability to learn more discriminative and informative representations for iris recognition.

\vspace{3px}
\noindent\textbf{Better leverage the unique characteristics of the iris texture:} Standard real-valued networks are designed to learn the appearance of objects through consistent shapes, edges, or other semantic structures \cite{NetworkViz}. However, the fact that the iris texture is stochastic \cite{IrisComplexity} with no consistent shapes, edges or stromal morphology would make real-valued networks struggle to learn any meaningful semantic structures from the iris texture and unable to realize the full potential of automatic feature learning. 

From a mathematical perspective, there are two key distinctive properties of complex-valued networks that are highly desirable in iris recognition.
\begin{itemize}
    \item Sensitive to phase: Complex-valued networks retain and process complex-valued features, hence they are able to retain and recover phase information. This make them sensitive to phase structure \cite{CVNN}. In iris recognition, phase is more important than magnitude \cite{DaugmanInformationTheory}. Phase encoding has been the key to many classical handcrafted features \cite{DaugmanFirstPaper,IrisCodeAnalysis,irisPhaseDCT,irisPhaseDFT,IrisZM}; however, standard real-valued networks have no direct mechanism to retain this. Complex-valued networks, in contrast, naturally allow us to directly and explicitly retain phase components due to its operation in a complex space.
    \item Locally stationary stochastic processes: The authors of \cite{ComplexCNN_Theory} have provided a solid mathematical framework to prove that complex-valued networks learn representations that are invariant to scale, resolution and orientation variations. All three properties, \emph{i.e.} invariant to multi-scale, multi-resolution and multi-orientation, are highly desirable to deal with variations in the iris texture.
\end{itemize}

\vspace{3px}
\noindent\textbf{Correspondence with Gabor wavelets:} In
\cite{ComplexCNN_Theory}, the authors have provided a mathematical framework to prove that complex-valued networks can be viewed as data-driven nonlinear multiwavelet packets. Gabor wavelets, as the core of the IrisCode, can be easily approximated by a complex-valued network. As depicted in Fig.~\ref{fig:architecturecomparison}, the automatic complex feature learning block in the proposed complex-valued network can converge to the handcrafted features in the IrisCode and shrink a complex-valued network to the IrisCode or its other phase-based derivatives. This generalization capability, interestingly, allows us to validate the optimality of the handcrafted IrisCode. During the training process, the network weights are learned to best represent the iris features. If the IrisCode features are truly the best representation for the iris problem, the training should converge the proposed network to emulate those features and we should achieve accuracy on par with the IrisCode. However, as will be shown in Section 4, the better recognition results of the proposed network prove that handcrafted IrisCode representations are not always the optimal answer and a new discriminative and more informative feature representation can be deduced directly from the data by the proposed network. Standard real-valued networks do not obviously exhibit the same exact correspondence with data-driven wavelets \cite{ComplexCNN_Theory}, making them less effective in the iris feature learning.  


In summary, our contribution to develop complex-valued networks for iris recognition is driven by multiple factors. Firstly, our approach will provide a richer representational capacity through a complex representation which is data driven through automatic feature learning. Secondly, our approach is driven by the biological characteristics of the iris texture, which is stochastic in nature and lacking in consistent shapes, edges or semantic structures. Through provision of a network which is able to retain and recover phase information in the iris textural features, we can better exploit its intrinsic features for greater accuracy while making it invariant to scale, resolution and orientation variations. Thirdly, by explicitly encoding phase information, we can directly compare our approach with both phase-based handcrafted approaches (such as the IrisCode) along with standard real-valued networks to validate their optimality for the iris recognition setting.

\begin{figure*}
\centering
\includegraphics[width=2\columnwidth]{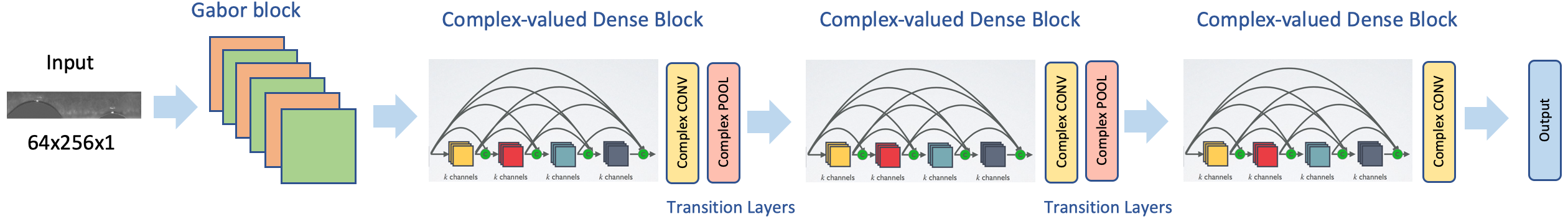}
\caption{ComplexIrisNet architecture. The normalized iris texture image is first fed through a Gabor Block, which generates the complex response to be subsequently processed by complex-valued operations. The following Dense Blocks and Transition Blocks are equipped with fully complex-valued operations, allowing them to process iris features in the complex-valued domain. The feature maps of the last CONV layer in the last Transition layer are used for feature representation.}
\label{fig:proposednet}
\end{figure*}

\section{The proposed Fully Complex-valued Iris Recognition Network}
\label{sec:proposed}
We propose ComplexIrisNet - a densely-connected fully-convolutional and complex-valued network architecture - for the iris recognition task. The proposed network is illustrated in Fig.~\ref{fig:proposednet}. There are three major characteristics to highlight for the proposed network:

\begin{itemize}
\item Fully complex-valued: All network operations and feature maps are complex-valued. This brings it in line with the nature of the iris feature representation since the interaction between real and imaginary components, such as phase, has long been acknowledged as important. 
\item Fully convolutional: The fully-connected layers at the end of a deep neural network typically do not preserve the spatial adjacency information present in the input image, unlike the earlier convolutional layers \cite{FCN}. Therefore, we do not employ fully-connected layers in our network architecture.
\item Densely connected: Each layer takes all preceding feature-maps as input. The compelling benefits of the densely connected mechanism are: it alleviates the vanishing-gradient problem, strengthens feature propagation, encourages feature re-use and substantially reduces the number of parameters \cite{DenseNet}. This architecture is the state-of-the-art in large-scale ImageNet visual challenges, with the dual benefit of multiple skip connections of the ResNet architecture \cite{ResNet} and network-in-network connections of the inception architecture \cite{Inceptionv4}. 
\end{itemize}



The architecture of the proposed network and complex-valued operations are discussed in Section~\ref{sec:ProposedArchitecture} and Section~\ref{sec:complexoperations}. Section~\ref{sec:TrainingInference} discusses the changes required to train and infer a complex-valued network. 

\subsection{Network Architecture}
\label{sec:ProposedArchitecture}
A normalized iris image is the input for the proposed network. The input is first fed into a Gabor Block, which embeds the input iris images into the complex space. The complex representations are subsequently processed by multiple Dense Blocks, which are densely connected and fully convolutional. Each Dense Block is followed by a Transition Block, which further encourages interaction within the representation and promotes compactness of the feature map. Output from the last CONV layer is used as a feature representation for the input iris.

\noindent \textbf{Stage 1: Input}
A normalized iris image with a size of $64\times256$ is used as input for the network. This normalized iris image is generated in the pre-processing step. The pre-processing step is described in Section~\ref{sec:preprocessing}.
\vspace{1.5mm}

\noindent \textbf{Stage 2: Gabor Block} consists of a group of complex-valued convolutional filters. We explicitly initialize these filters with Gabor-like complex kernels to simulate a family of Gabor wavelet transform in iris recognition. 
\begin{equation}
g(x,y; \lambda, \theta, \psi, \delta, \gamma) = \exp(-\frac{x'^2+\gamma^2y'^2}{2\delta^2}) \exp(i(2\pi\frac{x'}{\lambda}+\psi))
\end{equation}
where $x'=x \cos(\theta)+y \sin(\theta)$ and $y'= -x \sin(\theta) + y \cos(\theta)$. Here, $\theta$ is the orientation of the normal to the parallel stripes of a Gabor function, $\lambda$ is the wavelength of the sinusoidal factor, $\delta$ is the sigma/standard deviation of the Gaussian envelop and $\gamma$ is the spatial aspect ratio and specifies the ellipticity of the support of the Gabor function \cite{Gabor,DaugmanFirstPaper}. The filters in this layer are initialized with different values of these five parameters to cover the family of the Gabor wavelet transform \cite{DaugmanFirstPaper,DaugmanInformationTheory}. With $M$ filters, the output will be $ H \times W \times (2M)$ with $M$ real and $M$ imaginary components. These $M$ Gabor-like filters will transform the input iris image into the complex space.


Some researchers have also tried to combine Gabor filters with CNNs. For example, Kwolek\cite{Face_GFCNN} extracted intrinsic facial Gabor features to feed to the subsequent CNNs. Sawar \emph{et al.} replaced a number of certain weight kernels of a CNN with Gabor filters to reduce the number of parameters and the model complexity \cite{GACNN}. Luan \emph{et al.} modulated convolutional filters via Gabor filter banks to improve the robustness to geometric transformation of CNNs \cite{GCN}. Compared to \cite{Face_GFCNN,GACNN,GCN}, we take a different approach by employing the entirety of the complex-valued outputs after applying the Gabor filters instead of merely the real-valued components. More importantly, the complex-valued Gabor response is subsequently processed by a fully complex-valued network to retain the complex-valued nature of the representation through the entire network. 

This Gabor Block has several interesting characteristics worth mentioning. Visualization of deep learning networks has shown that most networks learn edge detectors like the effect of Gabor filters in the first layer \cite{RepLearning}. Interestingly, Gabor filters have also been very effective in the early stages of the pipeline for encoding the iris textures before other encoding operations such as the phase-quadrant quantization are engaged in the generation of IrisCode \cite{Daugman07}. Hence, this block, besides its main purpose to embed the input into the complex domain for further processing, also models the prior in the classic handcrafted features. Even though the network may be arguably able to learn the first layer itself, explicitly steering the network architecture with a Gabor Block not only benefits in naturally embedding the input into the complex domain, but it also reduces the number of parameters to learn, which subsequently results in reducing the required amount of training data.

\vspace{1.5mm}

\noindent \textbf{Stage 3: Dense Block} consists of a number of complex-valued convolutional layers with dense connectivity. The dense connectivity within layers in these blocks enables the modeling of complicated relationships within the complex-valued feature maps. The real-valued dense blocks have been shown to be very effective in encoding visual details in the large-scale visual tasks \cite{DenseNet}. Our dense blocks are specifically equipped with the complex-valued operations which will be discussed in Section~\ref{sec:complexoperations}. 

\begin{figure}
\centering
\includegraphics[width=\columnwidth]{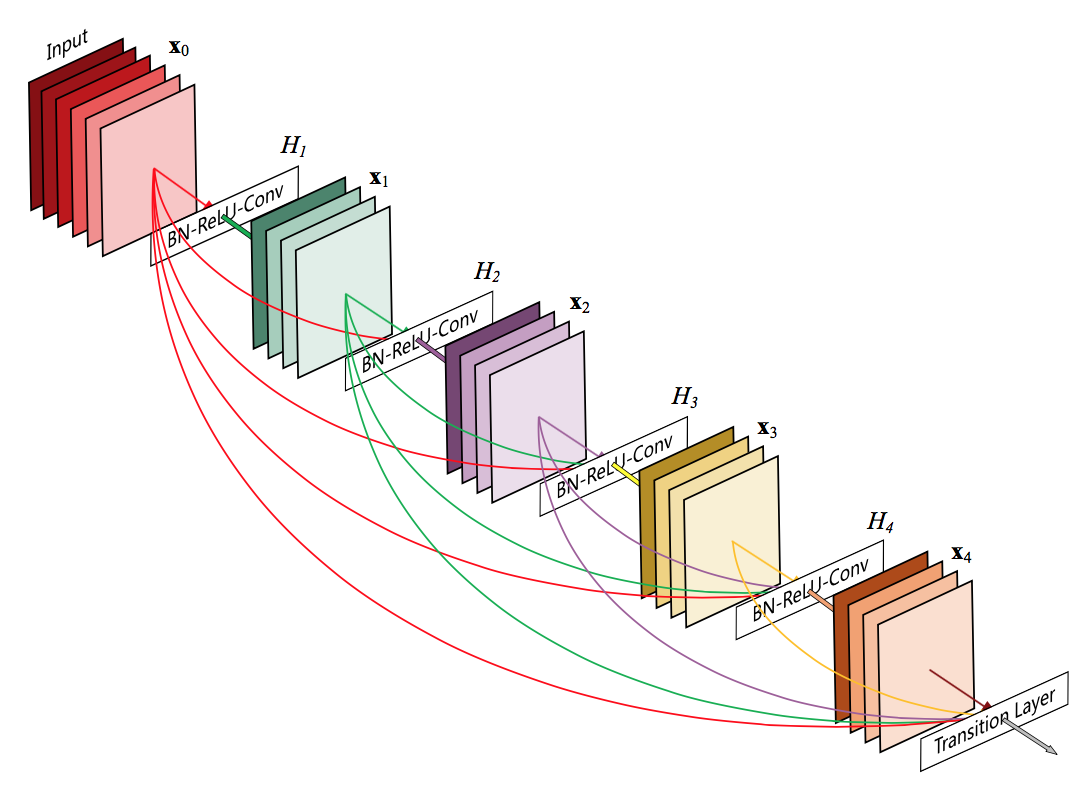}
\caption{Dense Block architecture. Each layer ($x_l$) in the block receives the feature maps of all preceding layers ($x_0,...,x_{l-1}$) as inputs. Six complex-valued network operations, $BN - ReLU - CONV (1\times1) - BN - ReLU - CONV (3\times3)$, are performed between two successive layers.}
\label{fig:denseblock}
\end{figure}

A dense block is illustrated in Fig.~\ref{fig:denseblock}. There are $N$ layers in each dense block. Each layer $x_i$ is fed through a combination of network operations as a function, $H_{i+1}$, to generate the next layer $x_{i+1}$. Six network operations are performed between layers: 
$BN - ReLU - CONV (1\times1) - BN - ReLU - CONV (3\times3),$ where, BN stands for complex-valued Batch Normalization, ReLU stands for complex-valued Rectified Linear Unit, and CONV stands for complex-valued Convolution. These operations will be discussed in Section~\ref{sec:complexoperations}. The dense connectivity is shown as the $l^{th}$ layer receives the feature maps of all preceding layers, $x_0,...,x_{l-1}$, as input, \emph{i.e.,}
\begin{equation}
x_l = H_l ([x_0,x_1,...,x_{l-1}]).
\end{equation}
The key benefit of the dense connectivity is allowing the signal to flow smoothly and mitigating the gradient vanishing and exploding effect, encouraging feature re-use, substantially reducing the number of parameters and thus enabling a deep visual representation \cite{DenseNet}.

Each Dense Block is followed by a Transition Block. A Transition Block contains one BN layer, one CONV layer, one ReLU layer and one $2\times2$ Pooling layer. The purpose of the Transition Blocks is to improve the compactness of the feature maps and to encourage interaction within the representation.

\vspace{1.5mm}
\noindent \textbf{Stage 4: Output}
The feature maps from the last CONV layer of the last Transition Block are used as the feature representation for the input iris. Depending on the number of Dense Blocks, the size of the output will vary. One example of the architecture of the proposed complex-valued deep iris network, ComplexIrisNet, with three Dense Blocks and three Transition Blocks is presented in Table~\ref{tab:Architecture}.

\begin{table}
\centering
\caption{Layer configuration of the ComplexIrisNet.}
\label{tab:Architecture}
\begin{tabular}{lllcr}
Blocks                                                      & \begin{tabular}[c]{@{}l@{}}Kernel\\ Size\end{tabular} & \begin{tabular}[c]{@{}l@{}}Output \\ Size\end{tabular} & \#Layers & \#Parameters  \\ 
\toprule
Input                                                       &                                                       & 64 x 256 x 1                                           &          &               \\ 
\hline
\begin{tabular}[c]{@{}l@{}}Gabor\\ Block\end{tabular}       & (7x7) x 40                                            & 32 x 128 x 40                                          & 4        & 1,960         \\ 
\hline
\begin{tabular}[c]{@{}l@{}}Dense\\ Block1\end{tabular}      & \begin{tabular}[c]{@{}l@{}}1x1\\ 3x3\end{tabular}     & 32 x 128 x 168                                         & 36       & 205,824       \\ 
\hline
\begin{tabular}[c]{@{}l@{}}Transition\\ Block1\end{tabular} & (1x1) x 32                                            & 16 x 64 x 32                                           & 4        & 5,376         \\ 
\hline
\begin{tabular}[c]{@{}l@{}}Dense\\ Block2\end{tabular}      & \begin{tabular}[c]{@{}l@{}}1x1\\ 3x3\end{tabular}     & 16 x 64 x 160                                          & 36       & 204,800       \\ 
\hline
\begin{tabular}[c]{@{}l@{}}Transition\\ Block2\end{tabular} & (1x1) x 70                                            & 8 x 32 x 20                                            & 4        & 3,200         \\ 
\toprule
Total                                                       &                                                       &                                                        & 84       & 421,160      
\end{tabular}
\end{table}


\subsection{Complex-valued Network Operations}
\label{sec:complexoperations}
The complex-valued network can be considered a generalization of its real-valued counterparts. While many concepts from real-valued networks can be generalized trivially, the lack of ordering of the complex field makes generalization of some concepts tricky \cite{ComplexCNN_MasterThesis}. Without total ordering, two general complex numbers are not comparable; specifically, the {\em min} and {\em max} operations are not defined. The ReLU layer, max pooling layer and the optimization problem itself all rely on these operators. This section discusses these non-trivial changes.

A complex number, $z \in \mathbb{C}$, consisting of a real component, $x \in \mathbb{R}$, and an imaginary component, $y \in \mathbb{R}$, is defined as follows,
\begin{equation}
\label{eq:complex}
z = x + iy \in \mathbb{C}, \quad \text{where} \; \sqrt{-1}=i. 
\end{equation}
The complex number can also be represented with a magnitude, $r \in \mathbb{R}$, and a phase, $\theta \in \mathbb{R}$, as $z = r e^{i\theta}$. A complex function is defined as, 
\begin{eqnarray}
f&:&\mathbb{C} \rightarrow \mathbb{C} \\
f(z) &=& u(z) +i v(z), \quad where \; u,v: \mathbb{R}^2 \rightarrow \mathbb{R}
\end{eqnarray}

\vspace{1.5mm}
\noindent \textbf{Complex CONV}
A convolution operation of a feature map, $W$, with a convolutional filter, $k$, is presented as $W*k$, where $*$ stands for the convolution operation. A complex-valued convolution operation of a complex-valued feature map, $W = A + iB$, with a complex-valued convolutional filter, $k = x + iy$, is also a complex-valued feature map, $W*k = (A+iB)*(x+iy) = (A*x-B*y) + i(B*x+A*y)$, since the convolution is distributive. 

\begin{figure}
\centering
\includegraphics[width=\columnwidth]{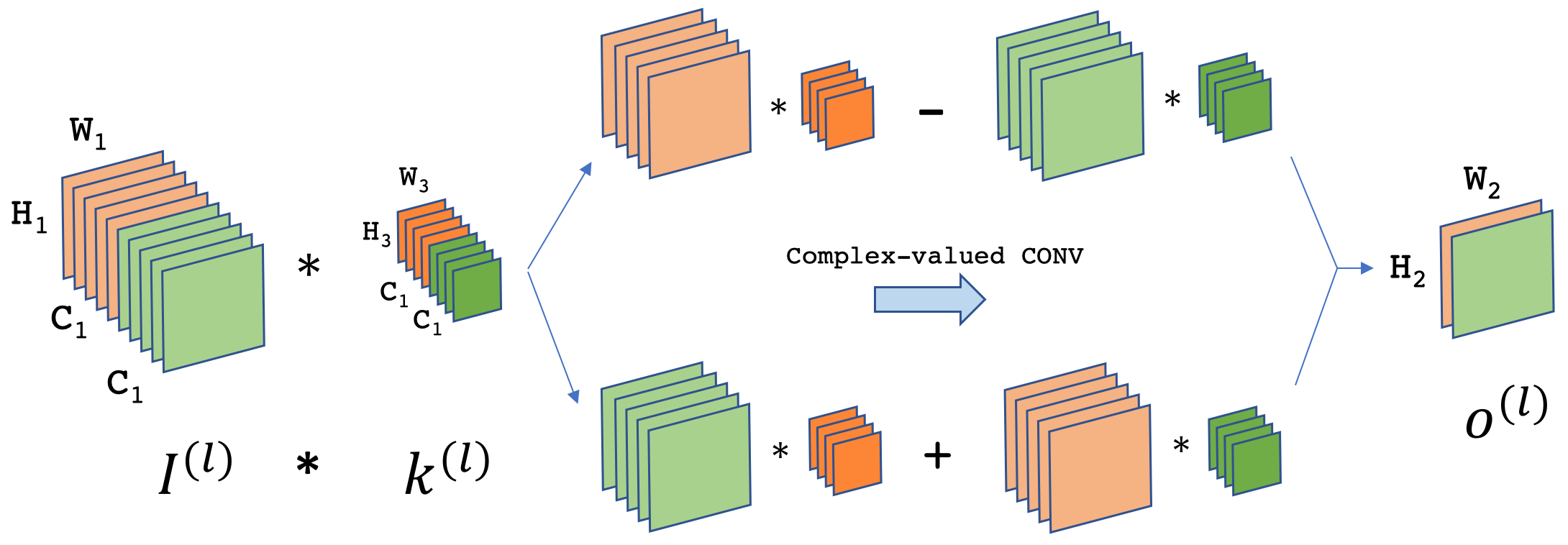}
\caption{Complex-valued convolution: the complex-valued input $I^{(l)}$ is convolved with a complex-valued kernel $k^{(l)}$ to output a complex-valued output $o^{(l)}$. For each convolutional layer, multiple kernels, $K^{(l)}$, are employed to generate the final output $O^{(l)}$.}
\label{fig:ComplexCONV}
\end{figure}

Mathematically, for one complex-valued convolutional layer $l$, the complex-valued input, $I^{(l)}$, which is the feature map output from the previous layer, is convolved with a set of complex-valued kernels, $K^{(l)}$, to generate a complex-valued output, $O^{(l)}$. 
\begin{itemize}
\item Input : $I^{(l)} \in \mathbb{C}^{W_1 \times H_1 \times (2C_1)}$ 
\item Complex-valued convolutional kernels: \\ $K^{(l)} \in \mathbb{C}^{W_3 \times H_3 \times (2C_1) \times (2C_2)}$
\item Output: $O^{(l)} = I^{(l)}*K^{(l)} \in \mathbb{C}^{W_2 \times H_2 \times (2C_2)}$ 
\end{itemize}
The kernel set has $C_2$ complex-valued kernels, each with a size of $W_3 \times H_3 \times (2C_1)$. The convolution of the input with one kernel is illustrated in Fig.~\ref{fig:ComplexCONV}. The output from convolving the input with $C_2$ kernels are concatenated to generate the final output $O^{(l)}$. There are two more parameters in convolution, stride, $S$, and zero-padding size, $P$. These parameters are similar to their real-valued counterpart. 

\begin{figure}
\centering
\includegraphics[width=0.8\columnwidth]{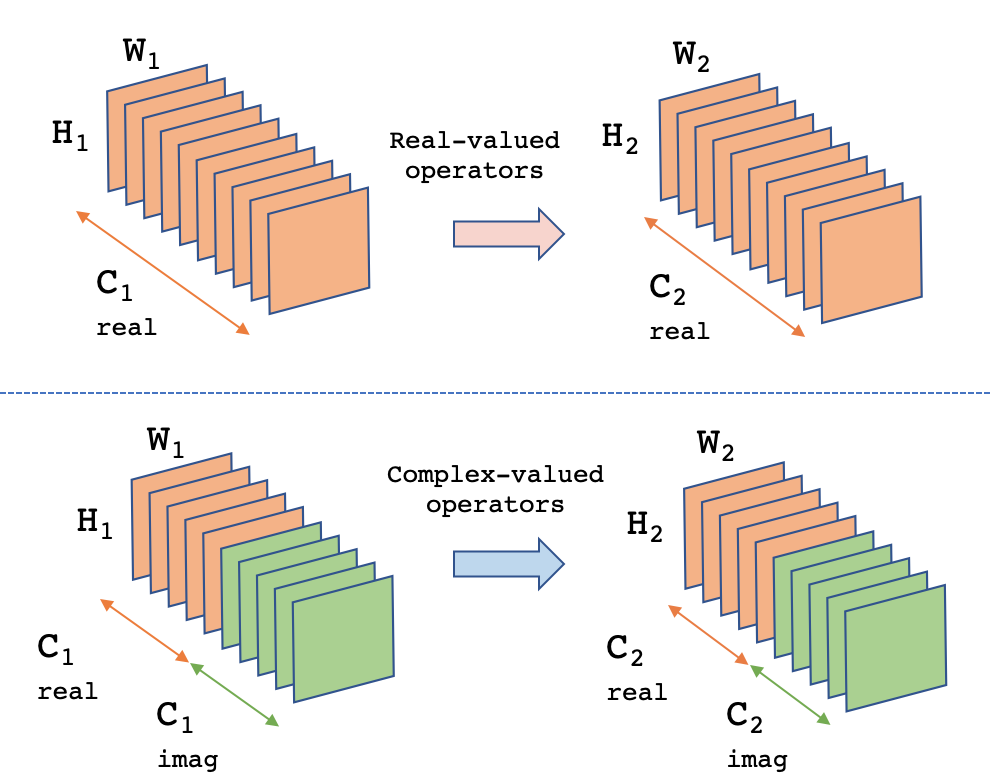}
\caption{Real-valued vs. Complex-valued operators and feature maps. Orange and green blocks denote real-valued and complex-valued feature maps, respectively. Pink and light blue arrows denote real- and complex-valued operations, respectively. While the volumetric shapes of the real- and complex-valued featuremaps look similar, the ways by which they are interpreted and calculated differ in nature.}
\label{fig:RealvsComplex}
\end{figure}

\vspace{1.5mm}
\noindent \textbf{Complex Activation}
Rectified Linear Unit (ReLU) is the most common activation function used. We generalize it to the complex domain as follows,
\begin{equation}
ReLU(z) = \left\{
                \begin{array}{ll}
                  z, \quad if \; arg(z) \in [0,\frac{\pi}{2}], \\
                  0, \quad otherwise.
                \end{array}
              \right.
\end{equation}

\noindent \textbf{Complex Pooling}
Spectral Pooling is the best candidate in comparison with others (Average Pooling, Max Pooling) since it can handle complex numbers \cite{SpectralPool}. In addition, the pooling is performed by truncating the representation in the frequency domain. This approach preserves considerably more information per parameter than other pooling strategies and enables flexibility in the choice of dimensionality of the pooling output. This representation also enables a new form of stochastic regularization by randomized modification of resolution. This has been shown to achieve competitive results on classification and approximation tasks, without using any dropout or max-pooling \cite{SpectralPool}. 

\vspace{1.5mm}
\noindent \textbf{Complex BN} Batch Normalization of a complex-valued input, $z$, is calculated as,
\begin{equation}
BN(z) = \gamma \tilde{z} + \beta,
\end{equation}
where $\tilde{z}$ is the normalization of $z$,
\begin{equation}
\tilde{z} = (V)^{-1/2} (z-E[z]).
\end{equation}
$V$ is the covariance matrix, $E$ is the mean of the input $z$.



\subsection{Training and inference in a complex-valued network}
\label{sec:TrainingInference}
The designed network is first trained to learn the weights that best encode the input image. Once trained, it can function as a feature extractor, which infers the feature representation from the normalized iris input image by performing a forward pass. This section highlights how the complex-valued domain shift changes the inference and training process.

\textbf{Inference}
The shift to the complex-valued domain requires to store both real-valued and imaginary-valued components in the feature maps. 
\begin{itemize}
\item A conventional real-valued layer performs a real-valued operation (\emph{i.e.,} real-valued conv, pooling, activation, etc.), to transform an input feature map of size $W_1 \times H_1 \times C_1$ to an output feature map of size $W_2 \times H_2 \times C_2$. 
\item A complex-valued layer performs a complex-valued operation (\emph{i.e.,} complex-valued conv, pooling, activation, etc. as discussed in Section~\ref{sec:complexoperations}) to transform an input feature map of size $W_1 \times H_1 \times (2C_1)$ to an output feature map of size $W_2 \times H_2 \times (2C_2)$. Half of the complex-valued input feature map, $W_1 \times H_1 \times (C_1)$, stores the real components and the other half, $W_1 \times H_1 \times (C_1)$, stores the imaginary components. 
\end{itemize}
Operations performed on each layer decide the relationship between the input and output feature maps as discussed in Section~\ref{sec:complexoperations}. Visualization of the feature map difference is illustrated in Fig.~\ref{fig:RealvsComplex}.

\textbf{Training}
The common way to train a network, \emph{i.e.,} to minimize the loss function, is by using an iterative steepest gradient descent algorithm (SGD). The SGD algorithm minimizes the loss function by iteratively updating the weights by moving step-wise in the direction of the loss function's steepest descent, which is opposite to its gradient. It iteratively performs: (1) a forward pass using the currently estimated weights to calculate the loss, then subsequently (2) a backward pass to update the weights based on how well the current weights have performed as assessed by the current loss value. The forward pass in the complex-valued domain has been discussed in the previous paragraph. The backward pass is the process to back-propagate the gradients of the loss function, $C$, through the network layers and using the chain rule to update the weights. 

In the conventional real-valued layers, the gradient is back-propagated from the layer $l_{i+1}$ to the layer $l_i$ as, 
\begin{equation}
\frac{\partial C}{\partial l_i} = \frac{\partial C}{\partial l_{i+1}} \frac{\partial l_{i+1}}{\partial l_i},
\end{equation}
where, $l_i$ is real-valued the feature map input at layer $i$-th and $l_{i+1}$ is real-valued the feature map input at layer $(i+1)$-th. 

In the complex-valued layers, the gradient of the loss at the layer $(i+1)$-th is calculated as,
\begin{equation}
\nabla_C(l_{i+1}) = \frac{\partial C}{\partial \Re l_{i+1}} + i \frac{\partial C}{\partial \Im l_{i+1}},
\end{equation}
where, $\Re l_{i}$ and $\Im l_{i}$ are the real and imaginary components of the complex-valued feature map at layer $i$. The chain rule is slightly different from the real-valued counterpart as,
\begin{eqnarray}
\nabla_C(l_{i}) &=& \frac{\partial C}{\partial \Re l_{i}} + i \frac{\partial C}{\partial \Im l_{i}} \\
&=& \bigg( \frac{\partial C}{\partial \Re l_{i+1}} \frac{{\partial \Re l_{i+1}}}{{\partial \Re l_{i}}} + \frac{\partial C}{\partial \Im l_{i+1}} \frac{{\partial \Im l_{i+1}}}{{\partial \Re l_{i}}} \bigg) \\ \nonumber
&+& i \bigg( \frac{\partial C}{\partial \Re l_{i+1}} \frac{{\partial \Re l_{i+1}}}{{\partial \Im l_{i}}} + \frac{\partial C}{\partial \Im l_{i+1}} \frac{{\partial \Im l_{i+1}}}{{\partial \Im l_{i}}} \bigg) \\
&=& \frac{\partial C}{\partial \Re l_{i+1}} \bigg( \frac{\partial \Re l_{i+1}}{\partial \Re l_i} + i \frac{\partial \Re l_{i+1}}{\partial \Im l_i} \bigg) \\ \nonumber
&+& \frac{\partial C}{\partial \Im l_{i+1}} \bigg( \frac{\partial \Im l_{i+1}}{\partial \Re l_i} + i \frac{\partial \Im l_{i+1}}{\partial \Im l_i} \bigg) \\
&=& \Re (\nabla_C (l_{i+1})) \bigg( \frac{\partial \Re l_{i+1}}{\partial \Re l_i} + i \frac{\partial \Re l_{i+1}}{\partial \Im l_i} \bigg) \\ \nonumber
&+& \Im (\nabla_C (l_{i+1})) \bigg( \frac{\partial \Im l_{i+1}}{\partial \Re l_i} + i \frac{\partial \Im l_{i+1}}{\partial \Im l_i} \bigg).
\end{eqnarray}
This chain rule allows the gradient of the loss to be back-propagated from the last layer through each layer in the network. The weights are updated accordingly.


\textbf{Loss function}
There are two types of loss functions to train a network: classification losses and pairwise losses. The majority of previous iris recognition approaches based on deep networks use classification losses trained over a set of known iris identities and then utilize the intermediate bottleneck layer as a representation scheme that extends beyond the set of identities used in training. The downsides of this approach are its indirectness and inefficiency \cite{FaceNet} while relying on the assumption of the generalizability of the bottleneck representation to new irises. Compared to classification losses, pairwise losses \emph{ directly reflect what we want to achieve, i.e., train the representation to correspond to iris similarity}: images of the same iris have small distances and images of different irises have larger distances. We leverage the recent success of a pairwise loss function called Extended Triplet loss as investigated in \cite{IrisFCN} to train the proposed network. Compared to \cite{IrisFCN} and \cite{FaceNet}, there is a fundamental change in optimizing the loss function due to shifting the weights from the real domain to the complex domain. 

To form a triplet, we need an anchor image, a positive image and a negative image. The positive image belongs to the same class with the anchor while the negative image belongs to a different class. Denoting the output vector of the network as $f$ (in our setting this would be the last convolutional layer), we can represent the output features for a particular triplet $i$ as $f_i^a;f_i^p;f_i^n$, denoting the output features for the anchor, positive and negative images, respectively. The goal of a triplet loss function is to make the distance between $f_i^a$ and $f_i^n$ (\emph{i.e.,} images from different classes) larger than the distance between $f_i^a$ and $f_i^p$ (\emph{i.e.,} images from the same class) by a minimum margin $\alpha$. The extended triple loss function is defined as,
\begin{equation}
ETL = \frac{1}{N} \sum_{i=1}^N \Big[ D(f_i^A,f_i^P) - D(f_i^A,f_i^N) + \alpha \Big],
\end{equation}
where, $D$ is the distance between two vectors. In the iris recognition case, $D$ between two feature maps, $f^1$ and $f^2$, has been designed to deal with eye rotations and segmentation masks as,
\begin{equation}
D(f^1,f^2) = \min_{-B \leq b \leq B} \Big\{ FD(f_b^1,f^2) \Big\},
\end{equation}
where, $b$ denotes that the feature map has been shifted horizontally by $b$ pixels; $B$ is the maximum of the shift allowed, \emph{i.e.,} 4 pixels in this work; and $FD$ is the Fractional Distance which takes the masks into consideration,
\begin{equation}
FD(f^1,f^2)=\frac{1}{|M|} \sum_{(x,y)\in M} \Big( f_{x,y}^1-f_{x,y}^2 \Big)^2.
\end{equation}




\section{Experimental results}
\label{sec:Experiments}

\begin{table*}
\caption{Statistics of three datasets, ND-CrossSensor-2013, CASIA-Iris-Thousand and UBIRIS.v2, used in this research.}
\centering
\label{tab:datasets}
\begin{tabular}{|l|l|l|l|l|l|l|l|l|} 
\hline
                    & \textbf{\# Subjects} & \textbf{\# Images} & \textbf{Distance}   & \textbf{Imager} & \begin{tabular}[c]{@{}l@{}}\textbf{Image}\\\textbf{Resolution}\end{tabular}& \begin{tabular}[c]{@{}l@{}}\textbf{Iris}~\\\textbf{Diameter}\end{tabular}& \textbf{Wavelength} & \begin{tabular}[c]{@{}l@{}}\textbf{Subject}~\\\textbf{Cooperation}\end{tabular}  \\ 
\hline
ND-CrossSensor-2013 & 676         & 111,564   & Close-up   & LG2200& 640x480& 200~& NIR& Highly                                                                                                                                                                \\ 
\hline
CASIA-Iris-Thousand & 1,000       & 20,000    & Close-up   & IKEMB-100& 640x480& 180& NIR& Highly                                                                                                                                                              \\ 
\hline
UBIRIS.v2           & 261         & 11,102    & 4-8 meters & CanonEOS 5D& 800x600& 180-80& Visible& Less                                                                                                                                                       \\
\hline
\end{tabular}
\end{table*}

We conducted our experiments on three public datasets: 
\begin{itemize}
\item \underline{ND-CrossSensor-Iris-2013 dataset \footnotemark:} is the largest publicly available iris dataset in the literature in terms of the number of images \cite{NDIRIS0405}. It contains 116,564 iris images captured by the LG2200 iris camera from 676 subjects. This is a super set of the ND-IRIS-0405 dataset with double the number of images (116,564 vs. 64,980) and the number of subjects (676 vs. 356). The number of images for each subject in the test set is between 11 and 312. After removing some falsely segmented samples, the test set contains 57,253 images in total, with 3,838,981 genuine pairs and 3,270,285,300 impostor pairs. \footnotetext{https://sites.google.com/a/nd.edu/public-cvrl/data-sets}

\item \underline{CASIA-Iris-Thousand dataset \footnotemark:} contains 20,000 iris images from 1,000 subjects, which were collected using the IKEMB-100 camera from IrisKing \cite{web:CASIA}. The test set, after removing some falsely segmented samples, contains 81,799 genuine pairs and 76,530,208 impostor pairs. \footnotetext{http://biometrics.idealtest.org}

\item \underline{UBIRIS.v2 iris dataset \footnotemark:} contains 11,102 iris images from 261 subjects with 10 images each subject. The images were captured under unconstrained conditions (at-a-distance, on-the-move and in the visible wavelength), with corresponding more realistic noise factors \cite{UBIRISv2}. The test set, after removing some falsely segmented samples, contains 9,629 genuine pairs and 4,317,548 impostor pairs. \footnotetext{http://iris.di.ubi.pt/ubiris2.html}
\end{itemize}

\begin{figure}
\centering
\includegraphics[width=\columnwidth]{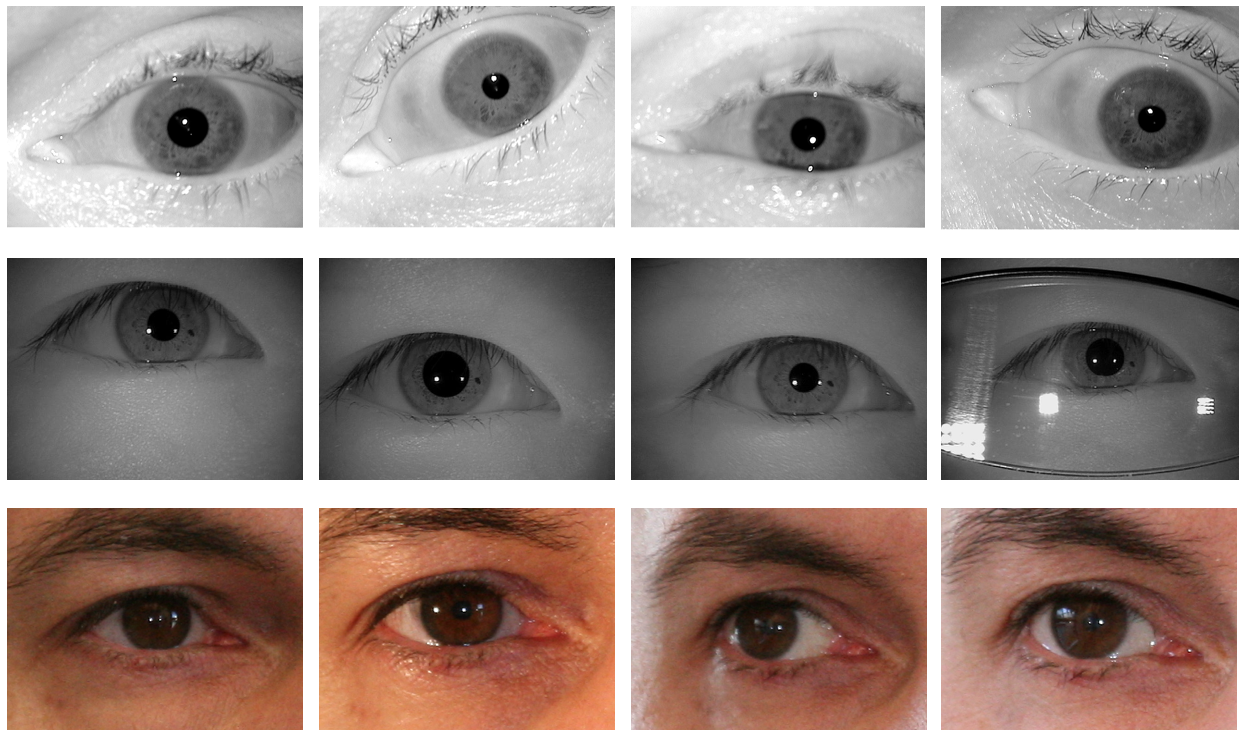}
\caption{Sample images from the ND-CrossSensor-2013 (first row), CASIA-Iris-Thousand (second row) and UBIRIS.v2 datasets (last row).}
\label{fig:samples}
\end{figure}

Experiments with these three datasets allow us to validate the performance of the ComplexIrisNet in diverse iris recognition scenarios, ranging from close-up distances in ND-CrossSensor-Iris-2013 and CASIA-Iris-Thousand to long distances in UBIRIS.v2; and from conventional near-infrared imaging in ND-CrossSensor-Iris-2013 and CASIA-Iris-Thousand to visible spectrum imaging in UBIRIS.v2. Sample images and statistics of the three datasets are depicted in Fig.~\ref{fig:samples} and summarized in Table~\ref{tab:datasets}.

\subsection{Performance metrics}
To report the performance of iris recognition approaches, we rely on Decision Error Trade-off (DET) curves and False Rejection Rates (FRRs). The DET curve is a graphical plot that illustrates the diagnostic ability of a classifier by calculating False Rejection Rate (FRR) against False Acceptance Rate (FAR) at various threshold settings. In this work, FRRs at FAR = $0.1\%$ are reported due to its popular adoption in the field. Equal Error Rate (EER) is the operating point where FAR is equal to FRR.

Due to the differences in the number of images, the ND-CrossSensor-Iris-2013 dataset is chosen for our intra-dataset experiments and the other two (CASIA and UBIRIS) are used for cross-dataset experiments. 

\textbf{Intra-dataset performance}
Similar to \cite{IrisFCN}, we train and test on disjoint identities. We use $80\%$ of the left eye images for training, $20\%$ of the left eye images for validation and all right eye images for testing. This guarantees that none of the classes in the test subset is in the train subset to avoid experimental bias. The training subset is used to train the ComplexIrisNet in an end-to-end manner to find the best weights. The validation subset is used to tune hyperparameters for the ablation study. The testing subset is used to report the intra-dataset performance. 

\textbf{Cross-dataset performance}
The proposed network is further investigated for generalization capability through training in one dataset and transferring the learned model to others. The pre-trained ComplexIrisNet, which has been pre-trained on the ND-CrossSensor-Iris-2013 dataset in the intra-dataset experiments, is subsequently cross-validated on two other datasets, CASIA-Iris-Thousand and UBIRIS.v2. We investigate two cross-dataset configurations: with and without fine-tuning. With fine-tuning, the pre-trained ComplexIrisNet is fine-tuned using all left eye images and is tested using all right eye images on the two datasets. Without fine-tuning, the pre-trained ComplexIrisNet is applied directly to all right eye images in the two datasets.

\subsection{Experimental setup} 
\label{sec:preprocessing}
We first pre-process the iris images via segmentation and normalization. The iris image is first segmented using two circles for the inner and outer boundaries of the iris region corresponding to the pupillary and limbus boundaries, respectively. We adopt a deep multi-task learning framework for joint iris segmentation and localization as in \cite{IrisParseNet}. The iris region from the raw Cartesian coordinates is then re-mapped to a dimensionless pseudo-polar coordinate, where the iris region is normalized to a fixed and rectangular size of $64\times256$ pixels \cite{Daugman07}. The corresponding noise mask is also normalized to facilitate matching in the later stages. 

It has to be stressed here that all algorithms discussed in the experiments were performed on the same test sets. As the preprocessing step may affect the overall performance, we ran the handcrafted, the state of the art and the proposed complex-valued algorithms on the same test sets for a fair comparison.

\begin{figure}
\centering
\includegraphics[width=\columnwidth]{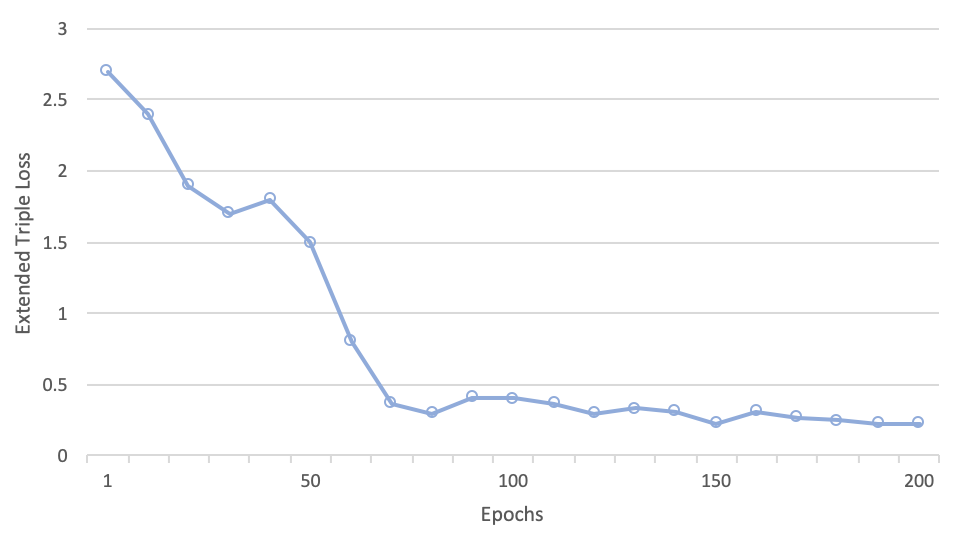}
\caption{Training Losses of the ComplexIrisNet on the ND-CrossSensor-Iris-2013 training subset.}
\label{fig:trainerror}
\end{figure}

The ComplexIrisNet is trained using the back-propagation algorithm with Stochastic Gradient Descent and with the Nesterov momentum set at $0.9$. The norm of gradients is clipped to $1$. The learning rate is initialized at $0.01$ for the first $10$ epochs, then set to $0.1$ from epoch $10$ to $100$, and then annealed by a factor of $10$ at epochs $130$ and $160$. The training error of the ComplexIrisNet is summarized in Fig.~\ref{fig:trainerror}. 

\begin{figure}
\centering
\includegraphics[width=\columnwidth]{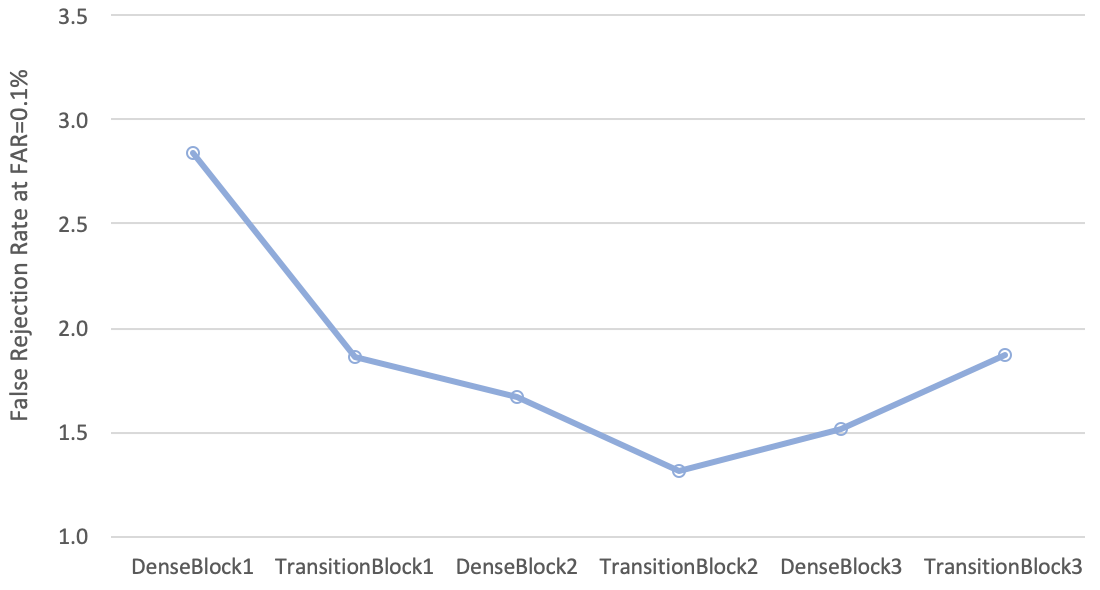}
\caption{Performance comparison of different layers of the ComplexIrisNet in terms of False Rejection Rate at $0.1\%$ False Acceptance Rate on the ND-CrossSensor-Iris-2013 validation subset.}
\label{fig:layerperformance}
\end{figure}

\subsection{Ablation study}
\subsubsection{Impact of dense architecture variants}
We first investigate how the architecture in terms of the number of dense blocks affects the performance. We vary this number from 1 to 3, since in the natural image classification setting, 3 dense blocks have shown to achieve state-of-the-art performance \cite{DenseNet}. Each dense block is followed by a transition block as explained in Section~\ref{sec:ProposedArchitecture}. Considering that the iris images are not as diverse as the natural images, a smaller-sized network may be able to achieve good results and avoid overfitting. Our experimental results show that a network with two dense blocks achieves the best performance during validation with a $1.31\%$ FRR (False Reject Rate) at a FAR (False Accept Rate) of $0.1\%$ on the ND-CrossSensor-Iris-2013 validation subset as shown in Fig.~\ref{fig:layerperformance}. Even a network with one dense block exhibits very good performance - $2.84\%$ FRR at FAR = $0.1\%$ - proving the high modeling capacity of the proposed complex-valued architecture in the iris recognition setting. From this point on, two dense blocks were used in our ComplexIrisNet unless otherwise indicated. 

\subsubsection{Impact of modeling complex-valued feature maps}
\label{sec:ComplexImpactAblation}
We further investigate the benefits of performing complex-valued operations within a deep network by comparing with variants of real-valued counterparts. We consider four real-valued representations. As shown in Equation~\ref{eq:complex},
$z = x + iy = re^{j\theta} \in \mathbb{C},$ where $x,y,r,\theta \in \mathbb{R}$ are the real part, the imaginary part, the magnitude and the phase of $z$, respectively. Four real-valued baselines are:
\begin{itemize}
\item 1. $r$ - Pure real network: Treat the input image purely with a real-valued network, there is no Gabor Block to generate the complex-valued response. This is similar to all other deep iris recognition networks in the literature.
\item 2. $(x,y)$ - Double-channel real network: Treat the complex-valued Gabor response as one real-valued network.
\item 3. $(x),(y)$ - 2 separate networks (Real and Imaginary): Treat the complex-valued Gabor response as two real-valued networks: one processes the real response $x$, and the other processes the imaginary response $y$.
\item 4. $(r),(\theta)$ - 2 separate networks (Magnitude and Phase): Treat the complex-valued Gabor response as two real-valued networks: one processes the magnitude response $r$, and the other processes the phase response $\theta$.
\end{itemize}

\begin{figure}
\centering
\includegraphics[width=\columnwidth]{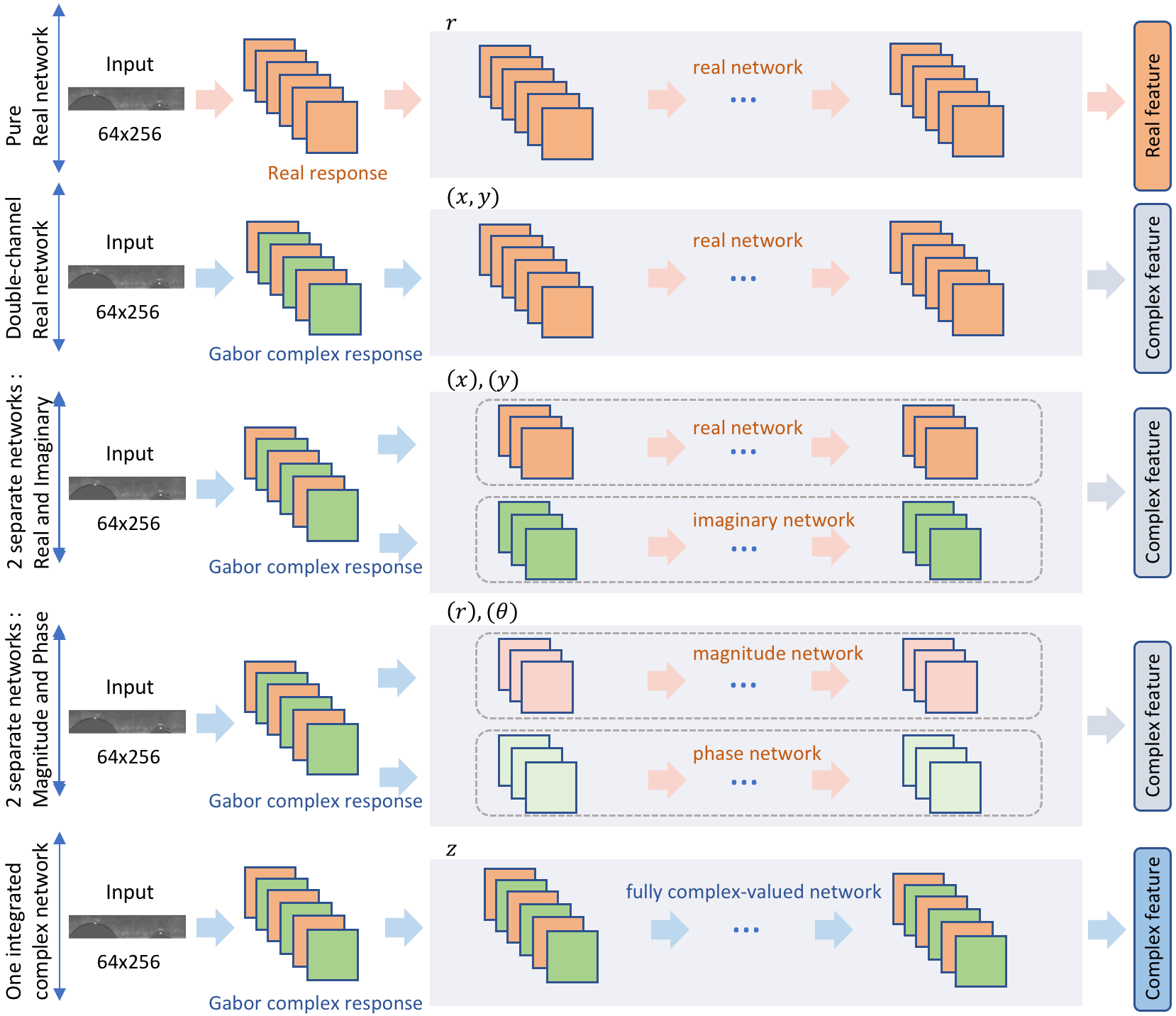}
\caption{Architecture comparison of four real-valued baselines and the complex-valued iris network. The complex-valued iris network enables complex-valued feature maps and complex-valued operations through the whole network.}
\label{fig:compleximpact}
\end{figure}



\begin{figure}
\centering
\includegraphics[width=\columnwidth]{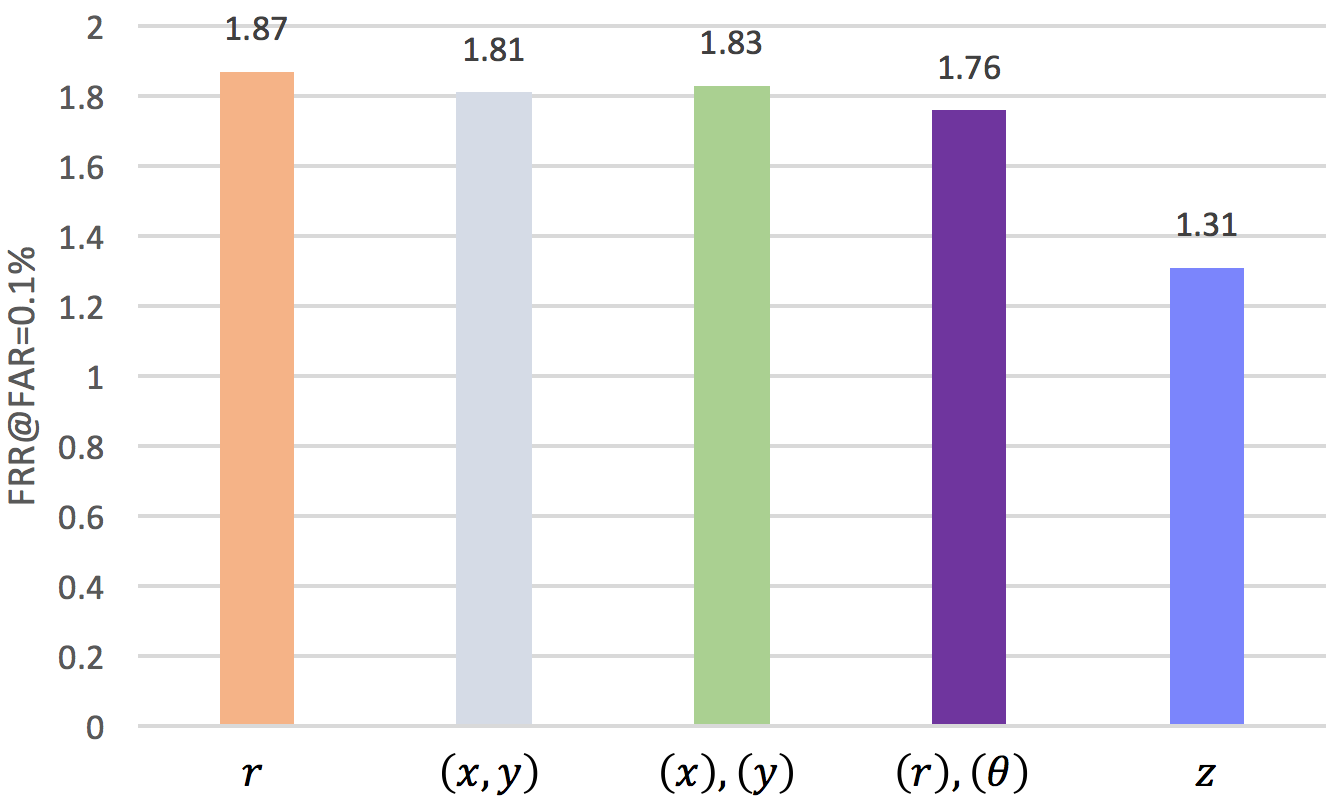}
\caption{Performance comparison of four real-valued network baselines and the proposed ComplexIrisNet on the ND-CrossSensor-Iris-2013 validation subset. This illustrates the benefits of performing complex-valued operations, which better capture the geometry and explicitly retain the phase information, which justifies its superior accuracy compared to the real-valued counterparts.}
\label{fig:realbaselines}
\end{figure}

Fig.~\ref{fig:compleximpact} explains the architectural difference between the four real-valued baselines and the proposed complex-valued iris network. It is clear that the real-valued networks, even with the same configurations (the number of layers, the number of filters per layer and the size of feature maps), would struggle to capture the phase and the intrinsic geometry of complex-valued data. This has been reinforced in terms of performance in Fig.~\ref{fig:realbaselines}.

The comparison illustrates the benefit of shifting the network operations to the complex domain. We hypothesize that the standard real-valued networks fail to fully capture the intrinsic geometric property of complex-valued data. Real-valued operations discard some of the geometric details and do not necessarily maintain the phase and spectral information of the data as it moves through the network. The complex-valued operations in our complex-valued network better capture the geometry and explicitly retain the phase information, which justifies its superior accuracy compared to the real-valued counterparts.

\vspace{3px}
\noindent \textbf{Visualization}\\
We visualize and analyze the network's predictions to understand the learning process underlying the complex network. GradCAM \cite{GradCAM} and CNN Fixation \cite{CNNfixation} are two popular methods to visualize CNNs; however, they are not designed for such pairwise networks as our ComplexIrisNet. The work in \cite{VizSimNet,VisualMetricLearning} have proposed to decompose the last convolution activation to highlight the image regions that contribute the most to the overall matching score. We employ these frameworks to visualize and compare the last convolution activation of both complex-valued and real-valued iris networks. 

\begin{figure}[h]
\centering
\includegraphics[width=\columnwidth]{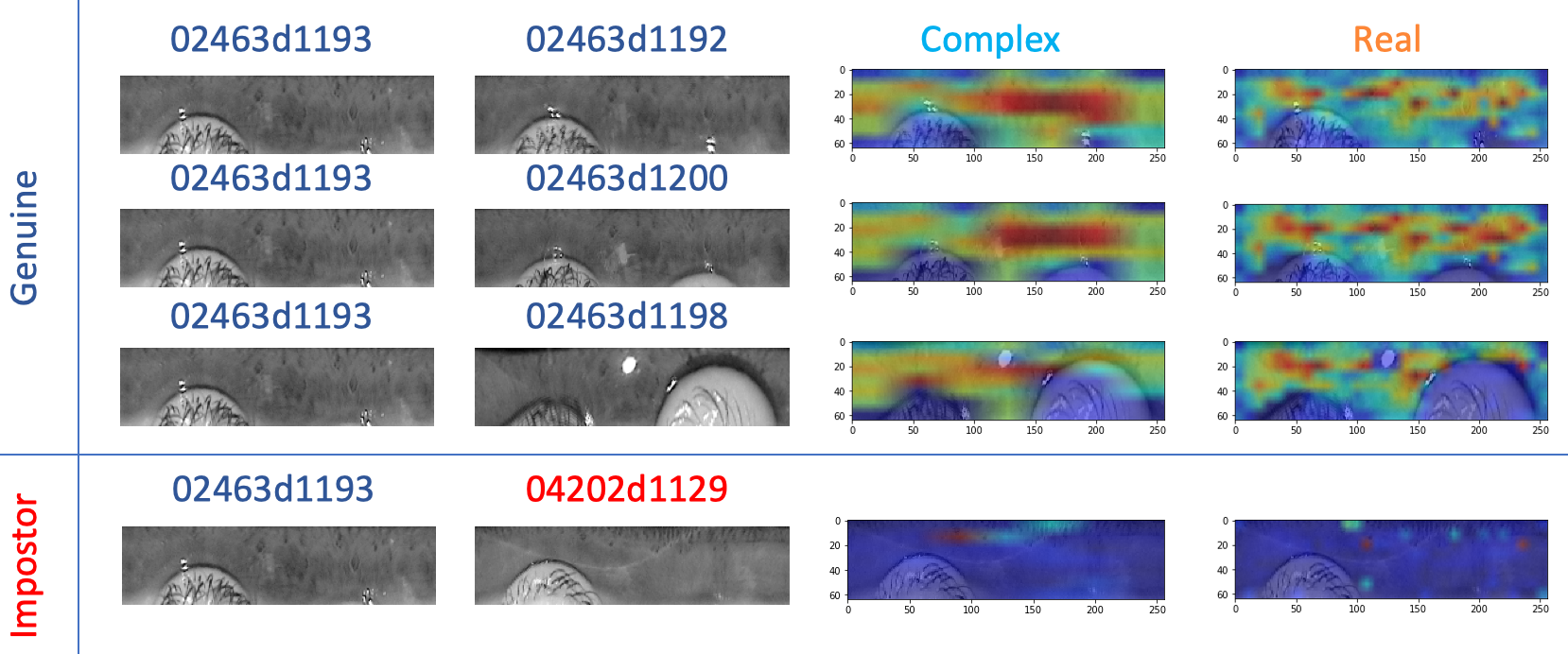}
\caption{Visualization of the activation decomposition of three genuine iris image pairs and one impostor iris image pair from the ND-CrossSensor-2013 dataset. The real iris networks tend to focus more on highly textured regions with significant spatial change, such as crypts. In contrast, the complex iris networks have a more diverse coverage, not just the regions with significant intensity change but also plainer texture regions, which exhibit less spatial change. The observation of this property in the network’s activations demonstrates the richer capacity of the complex network to capture both spatial and phase details compared to the real network counterparts.}
\label{fig:viz}
\end{figure}

Fig.~\ref{fig:viz} visualizes the activation decomposition of three genuine pairs and one impostor pair for the subject ID 02463 from the ND-CrossSensor-2013 dataset. The first two columns, respectively, depict the gallery and probe image pairs. The third and fourth columns, respectively, illustrate the heatmap overlays from the complex iris network and the real iris network. The heatmap overlays show the relative spatial contribution of each image region to the overall matching score. Hot colors denote high similarity while cold colors denote lower similarity. There are two observations from this figure to reflect on:
\begin{itemize}
    \item For real-valued iris network: The real networks tend to focus on the regions where the texture shows significant intensity change, which is demonstrated when many hot color regions in the real heatmap overlays are located in the crypts of the iris texture. In object detection and classification, real-valued deep networks are known for their capability to learn object structure via a hierarchical approach where earlier layers detect the basic edges and corners while later layers progressively detect more complex shapes \cite{RepLearning}. Unfortunately, this capability is not significant in the iris recognition setting due to the stochastic nature of the iris texture \cite{IrisComplexity}. The fact that the iris texture in iris recognition has no consistent shapes, edges, or structure unlike in classical object detection and classification, causes real-valued networks to struggle to learn any meaningful semantics from the iris texture and is unable to realize the full potential of automatic feature learning in the iris recognition setting. 

    \item For the complex iris network: The complex networks tend to have a more diverse coverage across the entire iris image, not just the regions with significant intensity change. They also focus on plainer texture regions, which visually seem to exhibit less spatial change. The observation of this property in the network’s activations demonstrates the richer capacity of the complex network to capture both spatial and phase details compared to its real network counterpart. In contrast to the spatial focus of the real iris networks, the complex iris networks also focus on phase information embedded within the image, making them more suitable for the iris recognition task. This highlights the fundamental value of the proposed complex-valued iris networks compared to the real-valued iris networks which is their ability to capture richer information (spatial and phase vs. spatial only). This ability allows complex iris networks to: (1) perform better genuine pair matching: better focus on the texture regions with less intensity value changes to recognize whether they are identical vs. real networks which only focus on significant intensity change regions; and (2) perform better impostor pair rejection: through an increased focus on both phase and spatial information to provide more clues to differentiate regions as shown in Fig.~\ref{fig:viz}. 
\end{itemize}






\begin{figure}
\centering
\includegraphics[width=0.9\columnwidth]{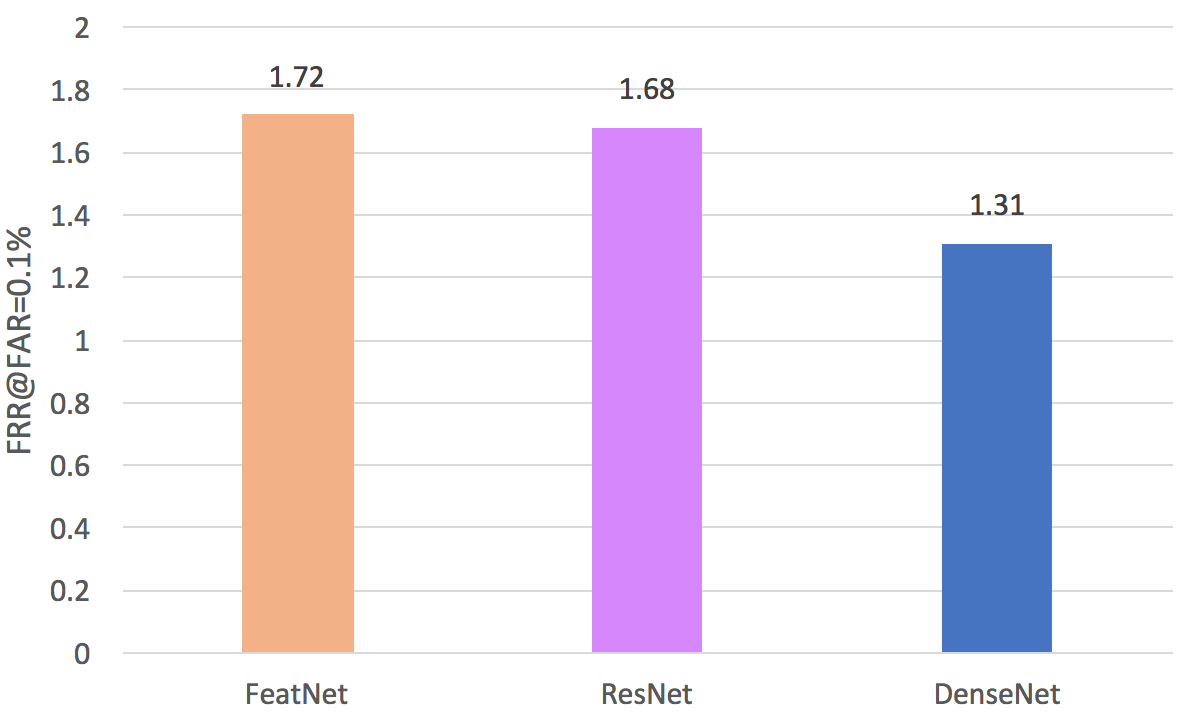}
\caption{Performance comparison of different backbone architecture choices for the complex-valued architecture on the ND-CrossSensor-Iris-2013 validation subset.}
\label{fig:deeparchitecturechoices}
\end{figure}




\begin{figure*}[!ht] 
  \subfloat[ND-CrossSensor-2013]{%
    \includegraphics[width=5.5cm]{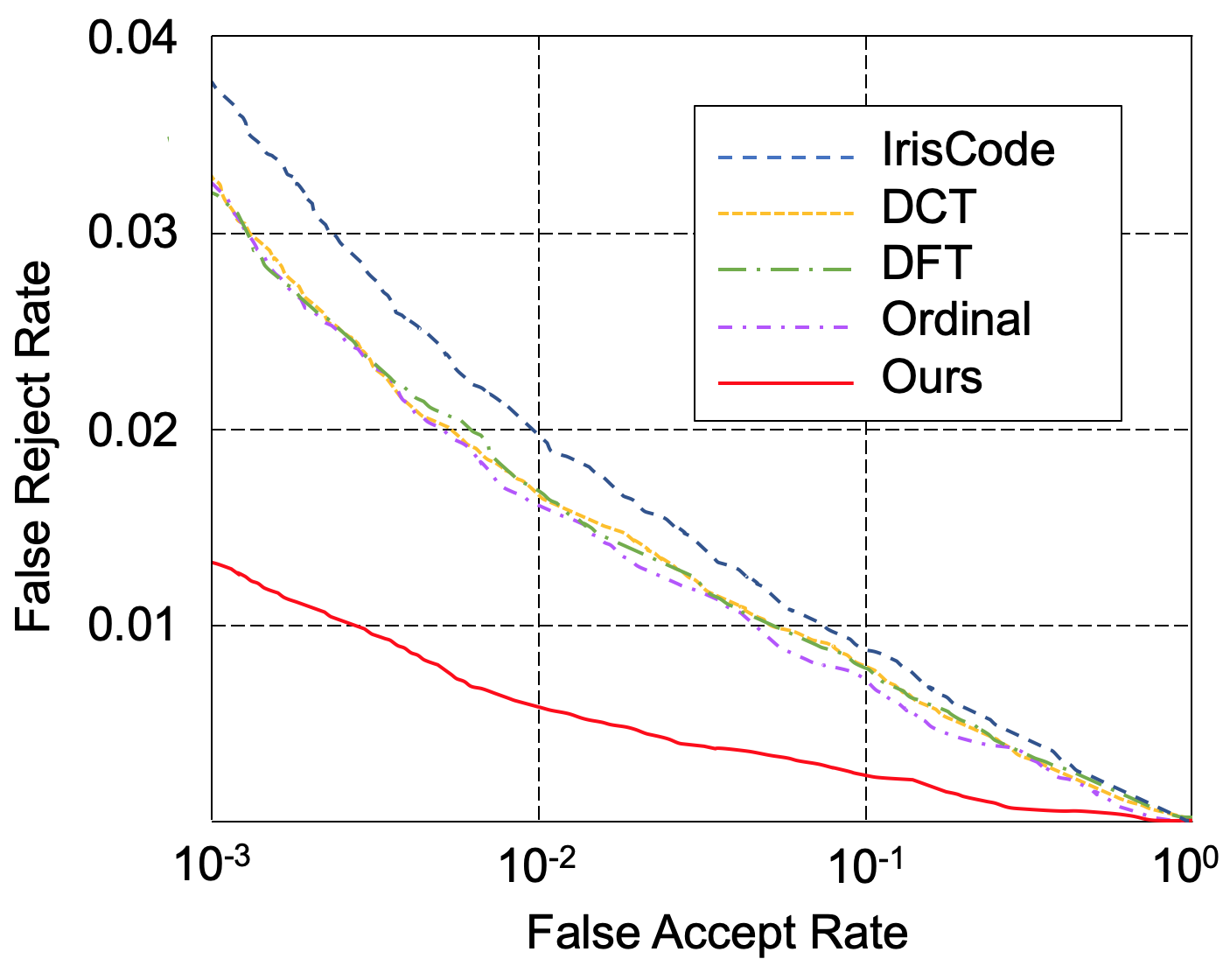} 
    \label{fig:comparisonHandcraftedA}
  } 
  \hfill 
  \subfloat[CASIA-Iris-Thousand]{%
    \includegraphics[width=5.5cm]{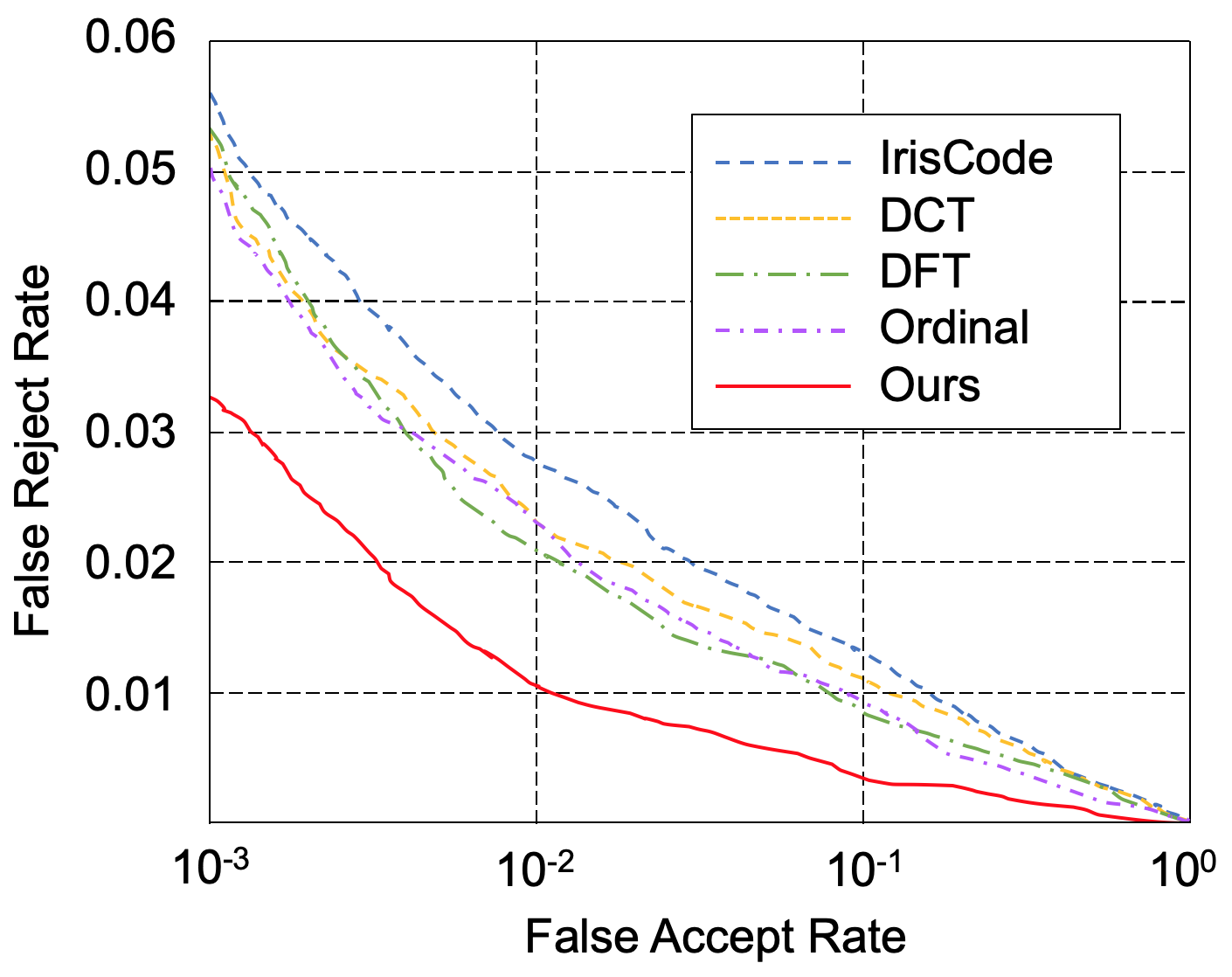} 
    \label{fig:comparisonHandcraftedB}
  } 
  \hfill 
  \subfloat[UBIRIS.v2]{%
    \includegraphics[width=5.5cm]{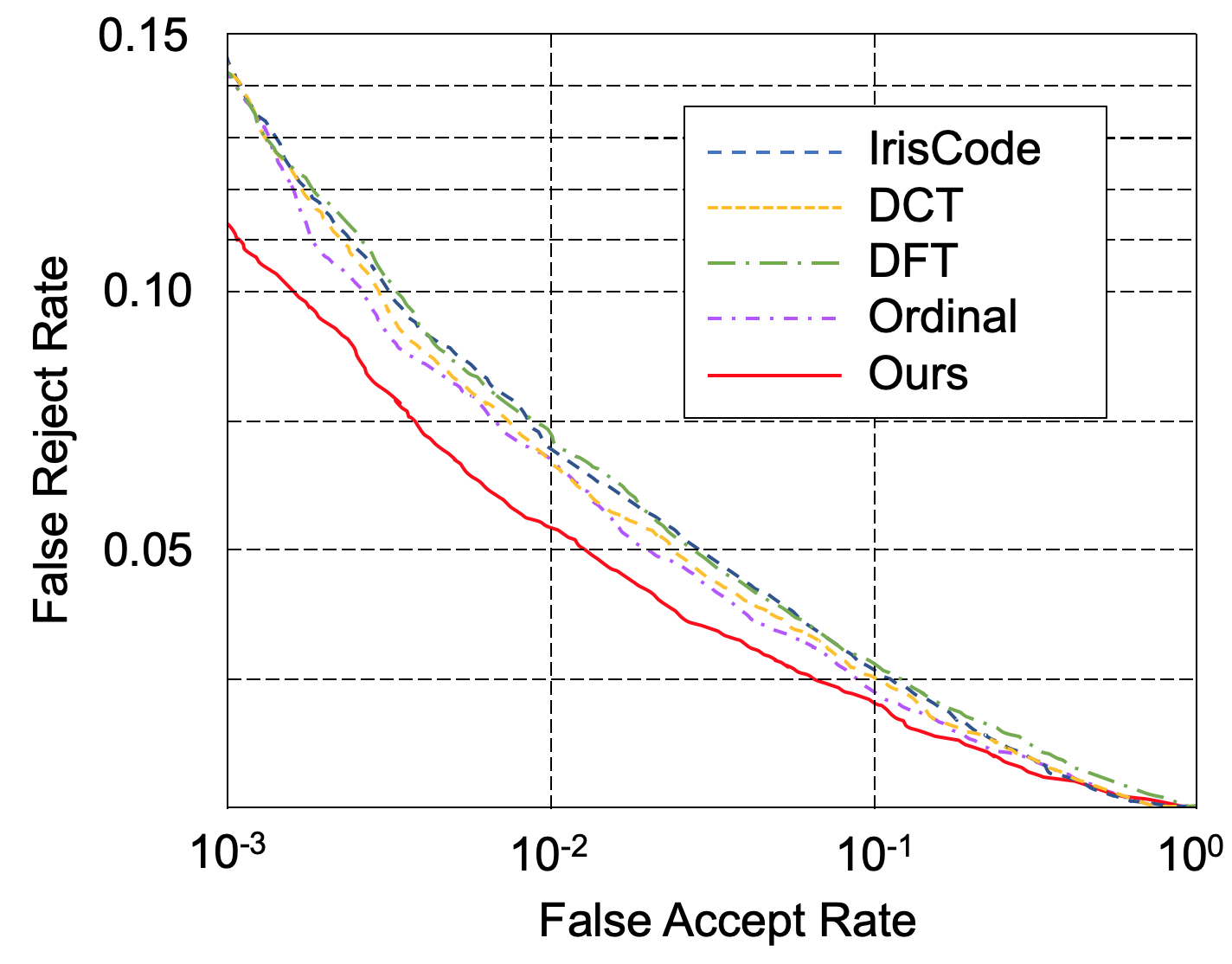} 
    \label{fig:comparisonHandcraftedC}
  } 
  \caption{DET curves for comparison of the proposed complex-valued data-driven features with other classic handcrafted feature representations on the test sets of three datasets. \emph{Best viewed in color.}} 
\label{fig:comparisonHandcrafted}
\end{figure*}

\subsubsection{Impact of backbone architecture choices}
\label{sec:ComplexArchitecture}
We now study how the choice of this backbone architecture compared to other architectures in the complex-valued domain. We retain the Gabor Block and swap the following dense blocks with one landmark architecture in general deep learning, \emph{viz.,} ResNet architecture \cite{ResNet}, and one landmark architecture in iris deep learning, \emph{viz.,} FeatNet architecture \cite{IrisFCN}. The FeatNet architecture is the state-of-the-art architecture in applying deep learning to iris recognition. FeatNet has three blocks of $CONV-TANH-POOL$. The feature maps from the $TANH$ activation layers are upsampled and stacked together before a fourth $CONV$ is applied. The ResNet architecture is the state-of-the-art (and among the most popular) architecture in general deep learning. Compared to architectures such as Inception \cite{Inceptionv4}, MobileNet \cite{MobileNet} and SENet \cite{SENet}, ResNet is chosen due to its notable generalization capacity together with the simplicity and consistency of its architecture. Details of FeatNet and ResNet architectures can be found in \cite{IrisFCN,ResNet}. We apply these two backbone architectures (FeatNet and ResNet) with the complex-valued operations as shown in Section~\ref{sec:complexoperations}. 

The effectiveness of the proposed architecture based on the dense blocks is validated by the performance on the validation subset as shown in Fig.~\ref{fig:deeparchitecturechoices}, achieving a FRR = $1.31\%$ compared to $1.68\%$ and $1.72\%$ at FAR = $0.1\%$ of ResNet and FeatNet, respectively.

\begin{table*}[]
\centering
\caption{Performance comparison with state-of-the-art handcrafted and deep learning iris recognition approaches in terms of False Rejection Rate at $0.1\%$ False Acceptance Rate and Equal Error Rate on the test sets of three datasets.}
\label{tab:comparison}
\begin{tabular}{l|m{1.1cm}|m{1.1cm}|m{1cm}|m{1cm}|m{1cm}|m{1cm}|m{1cm}|m{1cm}|m{1cm} m{1cm} }
\hline
                           & \multicolumn{2}{c|}{Intra-dataset}                                 & \multicolumn{8}{c}{Cross-dataset}         \\ \cline{2-11} 
                           & \multicolumn{2}{c|}{\multirow{2}{*}{\textbf{ND-CrossSensor-2013}}} & \multicolumn{4}{c|}{\textbf{CASIA-Iris-Thousand}}                       & \multicolumn{4}{c}{\textbf{UBIRIS.v2}}                                                                                                  \\ \cline{4-11} 
                           & \multicolumn{2}{c|}{}                                              & \multicolumn{2}{c|}{Fine-tuning}  & \multicolumn{2}{c|}{No fine-tuning} & \multicolumn{2}{c|}{Fine-tuning}                                              & \multicolumn{2}{c}{No fine-tuning}                     \\ \cline{2-11} 
                           & \textbf{FRR}                     & \textbf{EER}                    & \textbf{FRR}    & \textbf{EER}    & \textbf{FRR}      & \textbf{EER}    & \multicolumn{1}{l|}{\textbf{FRR}}     & \multicolumn{1}{l|}{\textbf{EER}}    & \multicolumn{1}{l|}{\textbf{FRR}}     & \textbf{EER}    \\ \hline
IrisCode                   & 3.43\%                           & 1.67\%                          & -               & -               & 5.33\%            & 3.37\%          & \multicolumn{1}{l|}{-}                & \multicolumn{1}{l|}{-}               & \multicolumn{1}{l|}{14.26\%}          & 8.30\%          \\ \hline
DCT \cite{irisPhaseDCT}                       & 3.21\%                           & 1.56\%                          & -               & -               & 5.18\%            & 3.29\%          & \multicolumn{1}{l|}{-}                & \multicolumn{1}{l|}{-}               & \multicolumn{1}{l|}{14.25\%}          & 8.26\%          \\ \hline
DFT \cite{irisPhaseDFT}                       & 3.18\%                           & 1.53\%                          & -               & -               & 5.19\%            & 3.30\%          & \multicolumn{1}{l|}{-}                & \multicolumn{1}{l|}{-}               & \multicolumn{1}{l|}{14.22\%}          & 8.23\%          \\ \hline
Ordinal \cite{irisOrdinalMeasures}                    & 3.17\%                           & 1.52\%                          & -               & -               & 5.01\%            & 3.21\%          & \multicolumn{1}{l|}{-}                & \multicolumn{1}{l|}{-}               & \multicolumn{1}{l|}{14.17\%}          & 8.07\%          \\ \hline
                           &                                  &                                 &                 &                 &                   &                 & \multicolumn{1}{l|}{}                 & \multicolumn{1}{l|}{}                & \multicolumn{1}{l|}{}                 &                 \\ \hline
DeepIris \cite{DeepIris}                   & 2.60\%                           & 1.29\%                          & 4.22\%          & 2.15\%          & 4.39\%            & 2.25\%          & \multicolumn{1}{l|}{13.21\%}          & \multicolumn{1}{l|}{7.13\%}          & \multicolumn{1}{l|}{13.67\%}          & 7.56\%          \\ \hline
FeatNet  \cite{IrisFCN,IrisFCNMaskRCNN}               & 1.79\%                           & 0.99\%                          & 3.96\%          & 1.92\%          & 4.01\%            & 2.03\%          & \multicolumn{1}{l|}{12.92\%}          & \multicolumn{1}{l|}{6.68\%}          & \multicolumn{1}{l|}{13.24\%}          & 7.03\%            \\ \hline
DRFNet   \cite{DRFNet}                & 1.77\%                           & 0.92\%                          & 3.91\%          & 1.89\%          & 4.05\%            & 2.00\%          & \multicolumn{1}{l|}{12.62\%}          & \multicolumn{1}{l|}{6.74\%}          & \multicolumn{1}{l|}{13.20\%}          & 6.88\%          \\ \hline
Complex (FeatNet backbone) & 1.45\%                           & 0.76\%                          & 3.66\%          & 1.75\%          & 3.72\%            & 1.81\%          & \multicolumn{1}{l|}{12.11\%}          & \multicolumn{1}{l|}{6.22\%}          & \multicolumn{1}{l|}{12.32\%}          & 6.14\%          \\ \hline
Complex (ResNet backbone)  & 1.41\%                           & 0.75\%                          & 3.67\%          & 1.76\%          & 3.70\%            & 1.79\%          & \multicolumn{1}{l|}{12.03\%}          & \multicolumn{1}{l|}{6.17\%}          & \multicolumn{1}{l|}{12.33\%}          & 6.18\%          \\ \hline
\textbf{ComplexIrisNet}    & \textbf{1.31\%}                  & \textbf{0.66\%}                 & \textbf{3.25\%} & \textbf{1.62\%} & \textbf{3.31\%}   & \textbf{1.64\%}   & \multicolumn{1}{l|}{\textbf{11.56\%}} & \multicolumn{1}{l|}{\textbf{6.01\%}} & \multicolumn{1}{l|}{\textbf{11.98\%}} & \textbf{6.15\%} \\ \hline
\end{tabular}
\end{table*}

\subsection{Comparison with Handcrafted Features}
We next compare the proposed method with handcrafted approaches. The major advantage of handcrafted features compared to deep learning features is fast computation and no requirement for training, which is beneficial for mobile and embedded applications. However, for applications such as large-scale identification systems running on high-end computing infrastructure, the recognition accuracy may be of more interest than the computation cost. The experiments in this section are designed to investigate how much performance can be leveraged by the proposed complex-valued data-driven features.

We compare with the classic IrisCode, its phase-based derivatives (DCT-based \cite{irisPhaseDCT} and DFT-based \cite{irisPhaseDFT}), and a state-of-the-art non-phase-based Ordinal features \cite{irisOrdinalMeasures}. As reviewed in Section 2, the IrisCode algorithm relied on non-linear encoding of the phase information extracted from the Gabor complex domain using a binarization scheme \cite{DaugmanFirstPaper,Daugman07}. We implemented the algorithm based on Daugman’s original papers \cite{DaugmanFirstPaper,Daugman07}, and used recent papers \cite{IrisFCN,IrisFCNMaskRCNN} to optimize the parameters for the highest potential performance and comparable experimental settings. We employ a bank of $M=40$ Gabor filters with 5 wavelengths and 8 orientations. For a fair comparison, similar to  \cite{irisOrdinalMeasures, IrisFCN}, we employ 72 ordinal filters. Other parameters are fine-tuned by a grid search for the best performance on the experimental datasets. 

For intra-dataset comparison, experiments are performed on the ND-CrossSensor-2013 dataset. The proposed complex iris networks with different backbones are first trained to learn a feature representation using the train set. Once the deep networks have been trained, they are employed to perform recognition on the test set. Since the handcrafted approaches do not require training, they are directly employed to perform recognition on the test set. The performance of the handcrafted features and complex iris networks with various backbones on the test set is illustrated in Fig.~\ref{fig:comparisonHandcraftedA} and the intra-dataset column of Table~\ref{tab:comparison}. Compared to an EER of $1.67\%$ for the classic IrisCode, the best complex iris network achieves an EER of $0.66\%$, which translates to a reduction in EER of $60\%$. 

For cross-dataset comparison, the complex iris networks, which have been trained in the intra-dataset experiments, are employed to perform recognition, with and without fine-tuning on the train sets of the CASIA-Iris-Thousand and UBIRIS.v2 datasets. Since the handcrafted approaches do not require training, these are again directly employed to perform recognition on the test sets. On the CASIA-Iris-Thousand dataset, the best complex iris network achieves EERs of $1.62\%$ and $1.64\%$ with and without fine-tuning, respectively. Compared to an EER of $3.37\%$ for the IrisCode, the complex iris network reduces EERs by $52\%$ and $50\%$ with and without fine-tuning, respectively. On the UBIRIS dataset, the best complex iris network achieves EERs of $6.01\%$ and $6.15\%$ with and without fine-tuning, respectively. Compared to an EER of $8.30\%$ for the IrisCode, the complex iris network reduces EERs by $26\%$ and $25\%$ with and without fine-tuning, respectively. Compared to the cross-dataset performance on the CASIA dataset, the performance gained from no-finetuning to finetuning on the UBIRIS dataset is less due to the domain shift from NIR to Visible in the latter.

The data shows that the features learned from the proposed ComplexIrisNets achieve consistently higher performance on all three datasets, reducing EERs by $25\%$ to $60\%$ in comparison with the traditional IrisCode. Similar performance gains over the other three handcrafted approaches are also observed. This not only shows that the feature representation automatically discovered by the ComplexIrisNet is highly discriminative achieving higher competitive performance, but it also illustrates the generalization capability of the discovered feature representation when tested on other datasets, both NIR (CASIA-Iris-Thousand) and Visible (UBIRIS.v2).

We note that the choice of a FAR at 0.1\% is to be consistent with the literature. This value has been widely used in deep learning papers for iris recognition such as \cite{IrisFCN,DeepIris,IrisFCNMaskRCNN}. Using this value enables us to directly compare our performance with theirs. We acknowledge that, as discussed in \cite{Daugman2021}, iris recognition has a large entropy and the IrisCode can generate very flat DET curves, \emph{i.e.} the FAR can be reduced by many orders of magnitude, even by factors of 10,000 to 100,000, while only paying a price of roughly doubling the FRR. This point has been confirmed in National Institute of Standards and Technology (NIST) reports \cite{NISTirexIII,NIST_IREXIX}. This is one of the advantages of iris recognition that makes IrisCode very successful.

\vspace{6px}
\noindent\textbf{VeriEye:} We also compare our method with the commercial product VeriEye SDK \footnote{https://www.neurotechnology.com/verieye.html} from NeuroTechnology. In 2020, Neurotechnology’s iris recognition algorithm was judged by NIST as the second most accurate among the IREX 10 participants. We obtained a copy of the VeriEye SDK and experimented on the ND dataset. We randomly selected different 100 irises, each having 10 images. There are a number of irises with less than 10 images in the dataset, for which we simply used all available images. In total, there are 973 images, generating 4,293 genuine image pairs and 468,585 impostor image pairs. We ran the VeriEye SDK, our home-brewed IrisCode and the ComplexIrisNet to compute matching scores. The results are presented in Table~\ref{tab:VeriEye}. The VeriEye SDK achieved a better EER compared to the home-brewed implementation of IrisCode. However, the ComplexIrisNet performs much better than VeriEye, thereby conveying the efficacy of the proposed method. 

\begin{table}
\centering
\caption{Comparison with the commercial VeriEye SDK.}
\begin{tabular}{|l|l|} 
\hline
               & EER     \\ 
\hline
Our IrisCode   & 1.66\%  \\ 
\hline
VeriEye        & 0.99\%  \\ 
\hline
ComplexIrisNet & 0.66\%  \\
\hline
\end{tabular}
\label{tab:VeriEye}
\end{table}

\subsection{Comparison with Deep learning Features}
We next compare with other deep-learning-based approaches in the literature. The state of the art of deep iris networks is based on pairwise losses with three state of the art approaches - DeepIris \cite{DeepIris}, FeatNet \cite{IrisFCN} and DRFNet \cite{DRFNet}. It is noteworthy that \cite{IrisFCNMaskRCNN} uses the same FeatNet as the backbone network to extract its iris feature representation. Since the source codes of these two approaches are not available, we re-implemented and optimized them to a comparable level of performance with those reported in the original papers.

We implemented DeepIris with 9 layers including one pairwise filter layer, one convolutional layer, two pooling layers, two normalization layers, two local layers and one fully connected layer \cite{DeepIris}. This network uses the normalized iris images with a size of $100\times100$ pixels; hence, in this experiment, we chose this output size for the segmentation process. 

We implemented the FeatNet with 4 convolutional layers, each followed by an activation layer ($Tanh$) and an average pooling layer as detailed in \cite{IrisFCN}. We implemented the Extended Triple Loss incorporating bit-shifting and masking \cite{IrisFCN}. This network is trained using the same pairwise loss scheme as the proposed ComplexIrisNet. One major difference is the architecture and the capacity to deal with complex-valued weights and feature maps. To purely compare the feature representation power, we used our own segmentation approach rather than their segmentation approach (MaskNet). Our implementation achieved comparable results to those reported in the paper (FRRs of $1.79\%$ vs. $1.78\%$ and same EER of $0.99\%$); the slight difference in performance may be due to the differences in the segmentation approach and dataset splitting scheme. We also implemented the improved version of FeatNet called DRFNet with dilated convolution and residual connections \cite{DRFNet}. DRFNet achieves slightly better performance in comparison with its plain version in FeatNet with FRR of $1.77\%$ and EER of $0.92\%$. 

It is noteworthy that the other two deep iris networks in the literature \cite{DeepIrisNet,OTS_CNN_Iris} use cross-entropy losses, which limits the applicability of the iris networks because the joint identities between the training and testing are required. The output of the intermediate layers from these networks can be extracted for use as feature vectors and some classifiers such as the Euclidean distance can be used to measure the (dis)similarity. However, their performance is inferior to the pairwise networks and are not included here. 

We also include the complex-valued networks with two different backbone architectures (FeatNet and ResNet) for comparison. 

\begin{figure}[!t] 
\centering
\includegraphics[width=0.9\columnwidth]{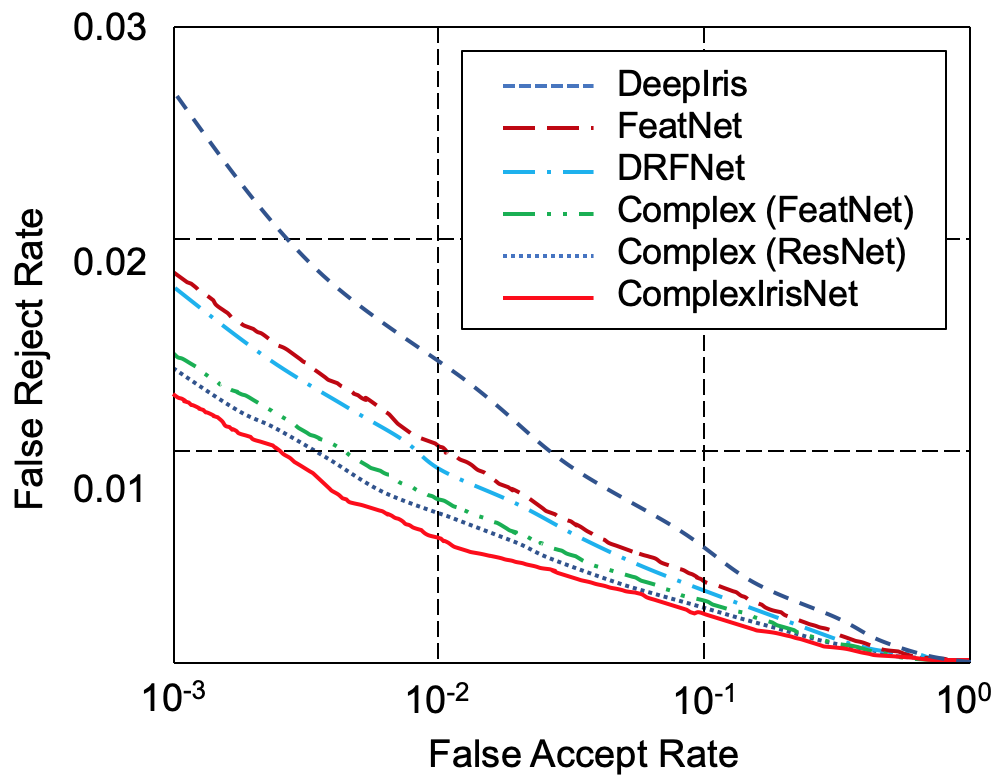}
\caption{DET curves for comparison with other deep learning feature representations on the test set of the ND-CrossSensor-2013 dataset. \emph{Best viewed in color.}} 
\label{fig:comparisonDL}
\end{figure}

For the intra-dataset scenario, experiments are performed on the ND-CrossSensor-2013 dataset. The proposed complex iris networks with various backbones and state of the art deep iris networks (DeepIris, FeatNet, DRFNet) are first trained on the same train set of the ND dataset. Once the networks have been trained, they are employed to perform recognition on the same test set of the ND dataset. Their performance is illustrated in Fig.~\ref{fig:comparisonDL} and the intra-dataset column of Table~\ref{tab:comparison}. The data shows that complex iris networks outperform all state of the art deep (real-valued) iris networks by large margins. The best complex iris network reduces the EER by $28\%$ - from $0.92\%$ for the best state of the art DRFNet \cite{DRFNet} to $0.66\%$ for the ComplexIrisNet.

For the cross-dataset scenario, the proposed complex iris networks and all state of the art deep (real-valued) iris networks, which have been pre-trained on the ND train set in the intra-dataset experiments, are employed to perform recognition, with and without fine-tuning, on the CASIA and UBIRIS datasets. The experimental results are shown in Fig.~\ref{fig:comparisonDL} and the bottom half of Table~\ref{tab:comparison}. The proposed ComplexIrisNet significantly reduces the EER by $18\%$ from $2.00\%$ to $1.64\%$ on the CASIA no fine-tuning experiments. Similar improvements have been shown in the UBIRIS dataset where the EER reduces from $6.88\%$ to $6.15\%$. This demonstrates the effectiveness of processing the weights and feature maps in the complex domain.

We also provide a complexity analysis and comparison as shown in Table~\ref{tab:complexity}. Three representative methods - DeepIris \cite{DeepIris}, FeaNet \cite{IrisFCN} and DRFNet \cite{DRFNet} - are compared to our method on the same machine. Our machine configuration is Intel(R) Core(TM) i7-8700K CPU @ 3.70GHz with 32GB DDR4 RAM, and NVIDIA GeForce RTX 2080 card with 8GB memory. It can be seen that the computational time of ComplexIrisNet is longer but acceptable (approximately 50 fps).

\begin{table}[]
\centering
\caption{Complexity Analysis}
\label{tab:complexity}
\begin{tabular}{|l|r|r|r|}
\hline
\textbf{Approach} & \textbf{\#Parameters} & \textbf{\begin{tabular}[c]{@{}l@{}}Model Size \\ (Byte)\end{tabular}} & \textbf{\begin{tabular}[c]{@{}l@{}}Feature \\ Extraction\end{tabular}} \\ \hline
DeepIris \cite{DeepIris}         & 55,420.0 K              & 289.0 M                      & 12.70 ms                     \\ \hline
FeatNet \cite{IrisFCN}          & 129.9 K               & 1.5 M                      & 8.93 ms                     \\ \hline
DRFNet \cite{DRFNet}           & 125.3 K               &    1.3 M                        & 8.12 ms                     \\ \hline
ComplexIrisNet              & 421.2 K               & 20.5 M                     & 21.08 ms                    \\ \hline
\end{tabular}
\end{table}

\section{Discussion and Conclusions}
\label{sec:Conclusion}

It is clear the iris stromal texture in iris recognition has no consistent shapes, edges, or structure unlike in classical object detection and classification task. This would make standard real-valued networks struggle to learn any meaningful semantic shapes from the iris texture during automatic feature learning and thereby not realizing the full potential of automatic feature learning in the iris recognition setting. 

Complex-valued networks have a solid mathematical and theoretical foundation that make them more suitable for iris recognition when compared to real-valued networks. 
\begin{itemize}
    \item Complex-valued networks explicitly model and retain \underline{phase information} through the whole network. Phase is very important in iris encoding as it has been shown in the success of many handcrafted features.
    \item Complex-valued networks are better at capturing features at \underline{multiple scales}, \underline{multiple frequencies} and \underline{multiple orientations} \cite{ComplexCNN_Theory}, which have the capacity to approximate such non-linear multiwavelet packets as Gabor wavelets in the classic IrisCode.
    \item Complex-valued networks allow \underline{automatic complex-valued feature learning}, which combines the strength of modern deep learning (automatic feature learning) and domain specific knowledge of the iris recognition field (complex filters). Compared to automatic real-valued feature learning in real-valued networks, automatic complex feature learning better suits the stochastic nature of the iris texture, leading to better feature learning.
    \item Complex-valued networks can be considered as a \underline{generalization of the classic IrisCode} due to their correspondence with Gabor wavelets. This allows us to understand the optimality of the classic IrisCode, since if it is optimal, the complex-valued networks will converge to its architecture and performance.
\end{itemize}

Despite these benefits, complex-valued networks have not been explored in the iris recognition literature. The key reason is due to the fact that the iris images themselves are not complex-valued, and so all existing deep iris networks operate on the real-valued intensity of iris images. 

We re-purpose the complex-valued response of Gabor filters to enable the development of new complex-valued networks to directly operate on iris images to translate all the above benefits to iris recognition. We also upgrade the dense connection and fully convolutional architecture to the complex-valued domain, which enables the proposed fully complex-valued network architecture to effectively and automatically learn a complex-valued feature representation for iris recognition. The proposed complex-valued iris network interestingly is a generalization of both the classic IrisCode and many state of the art deep iris networks. We show in our ablation study that complex-valued iris networks actually outperform their real-valued counterparts. More importantly, through visualizing the activation maps at the last convolutional layer, we show that while real-valued networks focus on capturing spatial detail, complex-valued networks are more sensitive to phase and are much better at capturing both spatial and phase information. The fundamental value of complex-valued networks compared to real-valued networks for iris recognition is the ability to capture richer detail (phase and spatial vs. spatial only) for better genuine pair matching and better impostor pair rejection. 

Through comprehensive comparison with handcrafted features and modern deep iris networks, we have shown that the feature representation discovered through the automatic complex feature learning process of the proposed method is more discriminative and informative than the feature representation discovered by the real-valued networks. The accuracy of the proposed complex-valued iris network surpasses the accuracy of the classic IrisCode (as implemented by us, judiciously guessing its parameter values), its phase-based derivatives and other state-of-the-art deep learning approaches by significant margins. The effectiveness and generalization capacity of the discovered feature representation has been demonstrated across three iris datasets under a variety of settings: near-infrared and visible spectral bands, near and far stand-off distances, and in strict and less cooperative scenarios.

Interestingly, in one of his original papers \cite{IrisCodeAsNetwork}, Daugman actually showed that IrisCode could be expressed as a two-layer CNN. The introduction of more layers and complex-valued operations, in theory, will lead to models with increased modeling capacity than classical approaches. The experiments conducted across multiple datasets have also, empirically, confirmed this observation.

This paper has presented another major step in the representation evolution in iris recognition, from classical handcrafted features to current deep learning with standard real-valued networks to tailored deep learning with complex-valued networks. With many benefits highly suitable for iris recognition, complex-valued networks deserve more attention and investigation from the iris recognition community to fully take advantage of automatic complex-valued feature learning. This work has taken the first step in this direction, and opens other opportunities for explicitly processing phase information in other domains where complex filters are routinely used for texture modeling.

{\small
\bibliographystyle{ieee}
\bibliography{ComplexIrisNet}
}

%

\begin{IEEEbiography}
    [{\includegraphics[width=1in,height=1.25in,clip,keepaspectratio]{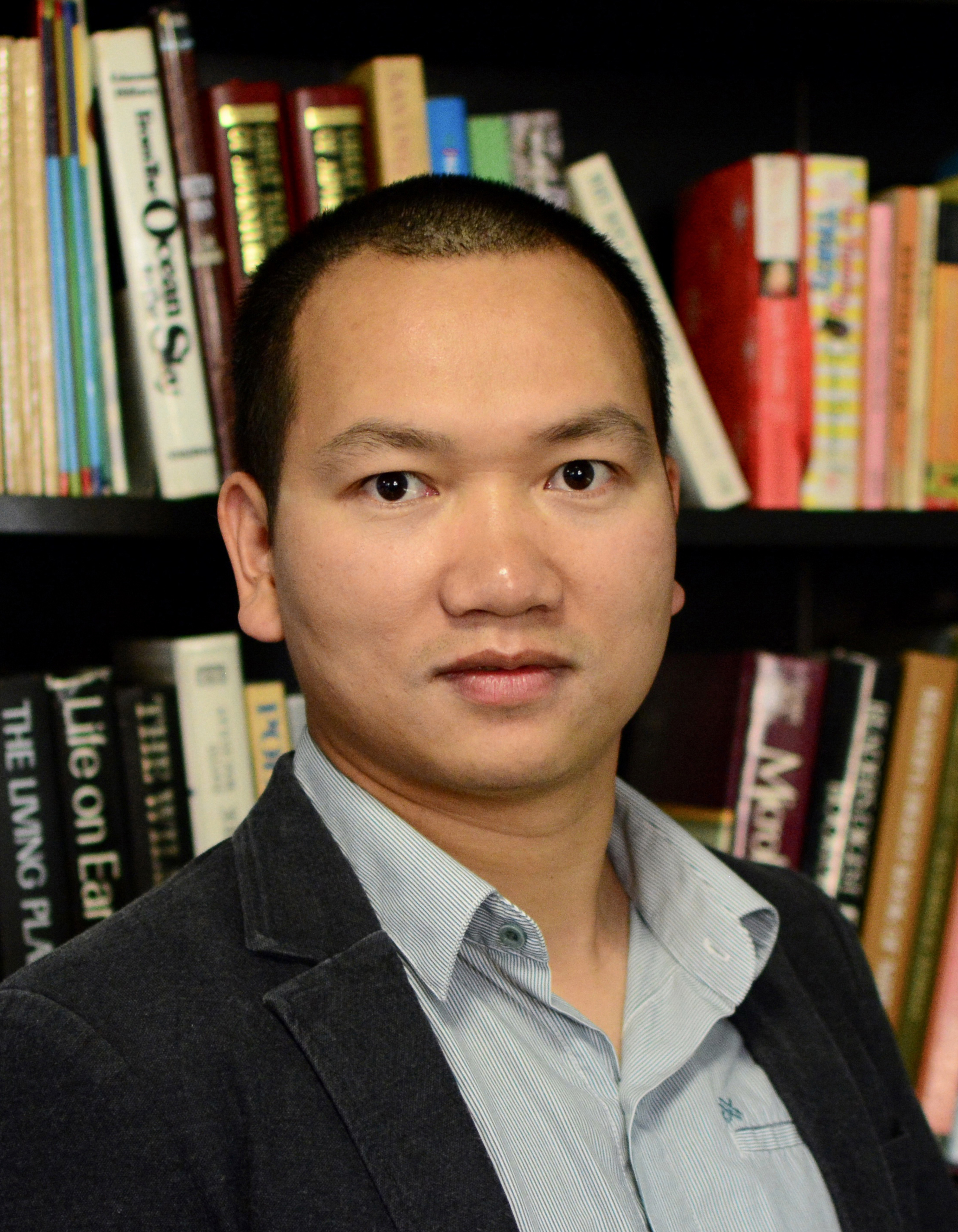}}]{Kien Nguyen}
is a Research Fellow at Queensland University of Technology. He has been conducting research in the area of iris recognition and biometrics for the last 10 years, and has published his research in high quality journals and conferences in the area. His research interests are in application of computer vision and deep learning techniques to the areas of biometrics, surveillance and scene understanding. He has been serving as an Associate Editor of the journal IEEE Access since 2016.
\end{IEEEbiography}

\begin{IEEEbiography}
    [{\includegraphics[width=1in,height=1.25in,clip,keepaspectratio]{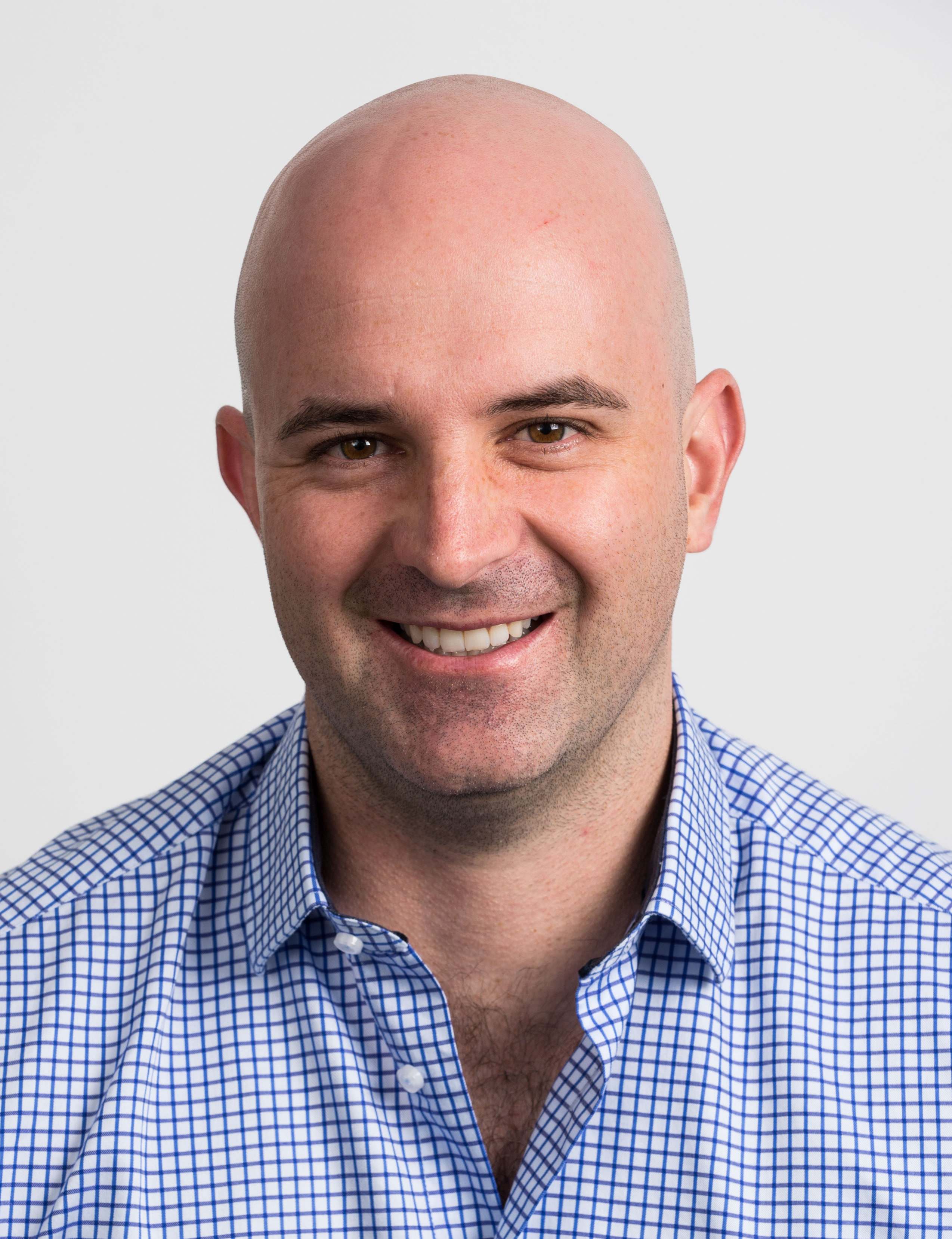}}]{Clinton Fookes} is a Professor in Vision and Signal Processing at the Queensland University of Technology. He holds a BEng (Aero/Av), an MBA, and a PhD in computer vision. He actively researches across computer vision, machine learning, signal processing and pattern recognition areas. He serves on the editorial boards for the IEEE Transactions on Image Processing, Pattern Recognition, and the IEEE Transactions on Information Forensics and Security. He is a Senior Member of the IEEE, an Australian Institute of Policy and Science Young Tall Poppy, an Australian Museum Eureka Prize winner, and a Senior Fulbright Scholar.

\end{IEEEbiography}

\begin{IEEEbiography}
    [{\includegraphics[width=1in,height=1.25in,clip,keepaspectratio]{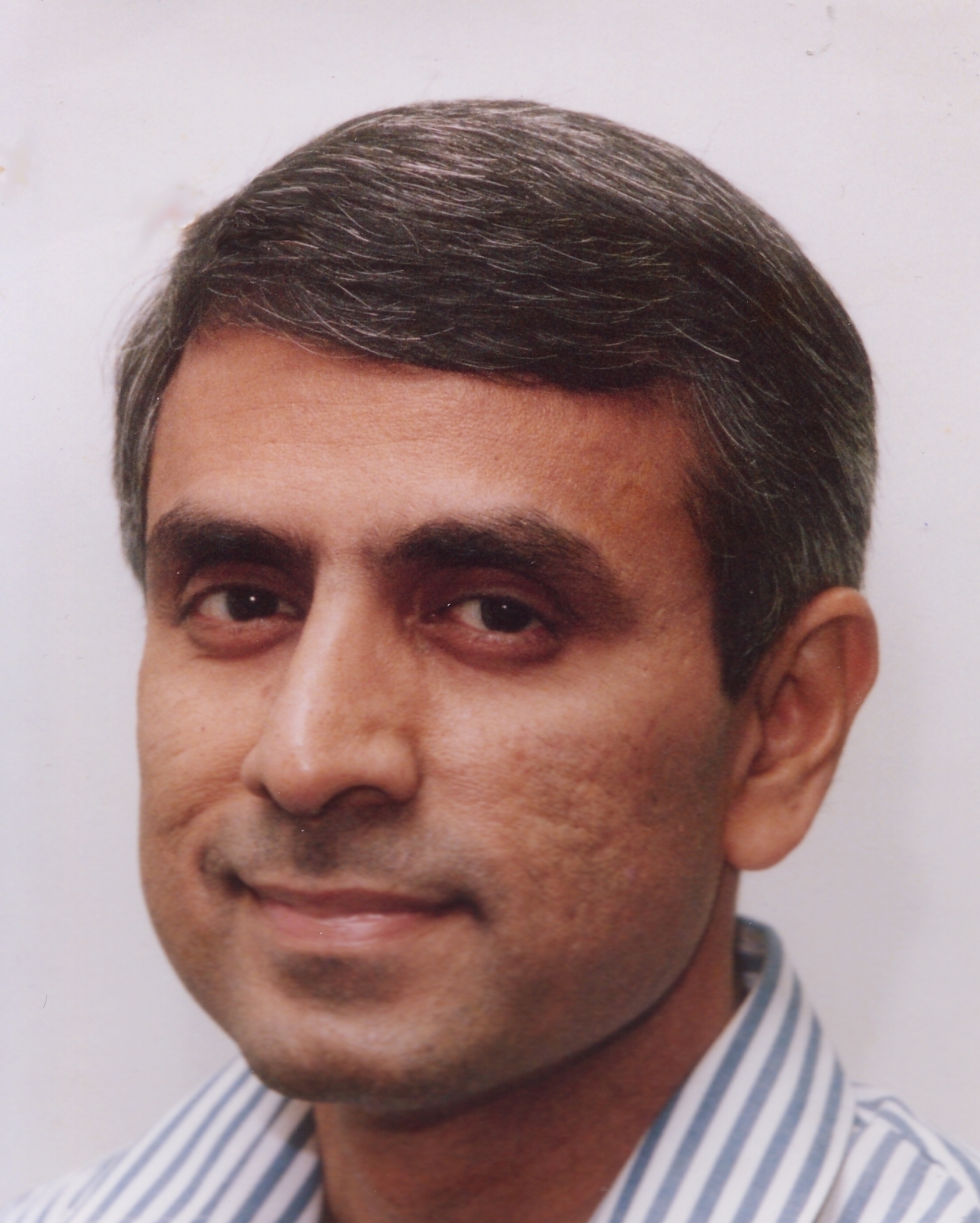}}]{Sridha Sridharan} obtained his MSc degree from the University of Manchester, UK and his PhD degree from University of New South Wales, Australia. He is currently a Professor at Queensland University of Technology (QUT) where he leads the research program in Signal Processing, Artificial Intelligence and Vision Technologies (SAIVT).
\end{IEEEbiography}

\begin{IEEEbiography}
    [{\includegraphics[width=1in,height=1.25in,clip,keepaspectratio]{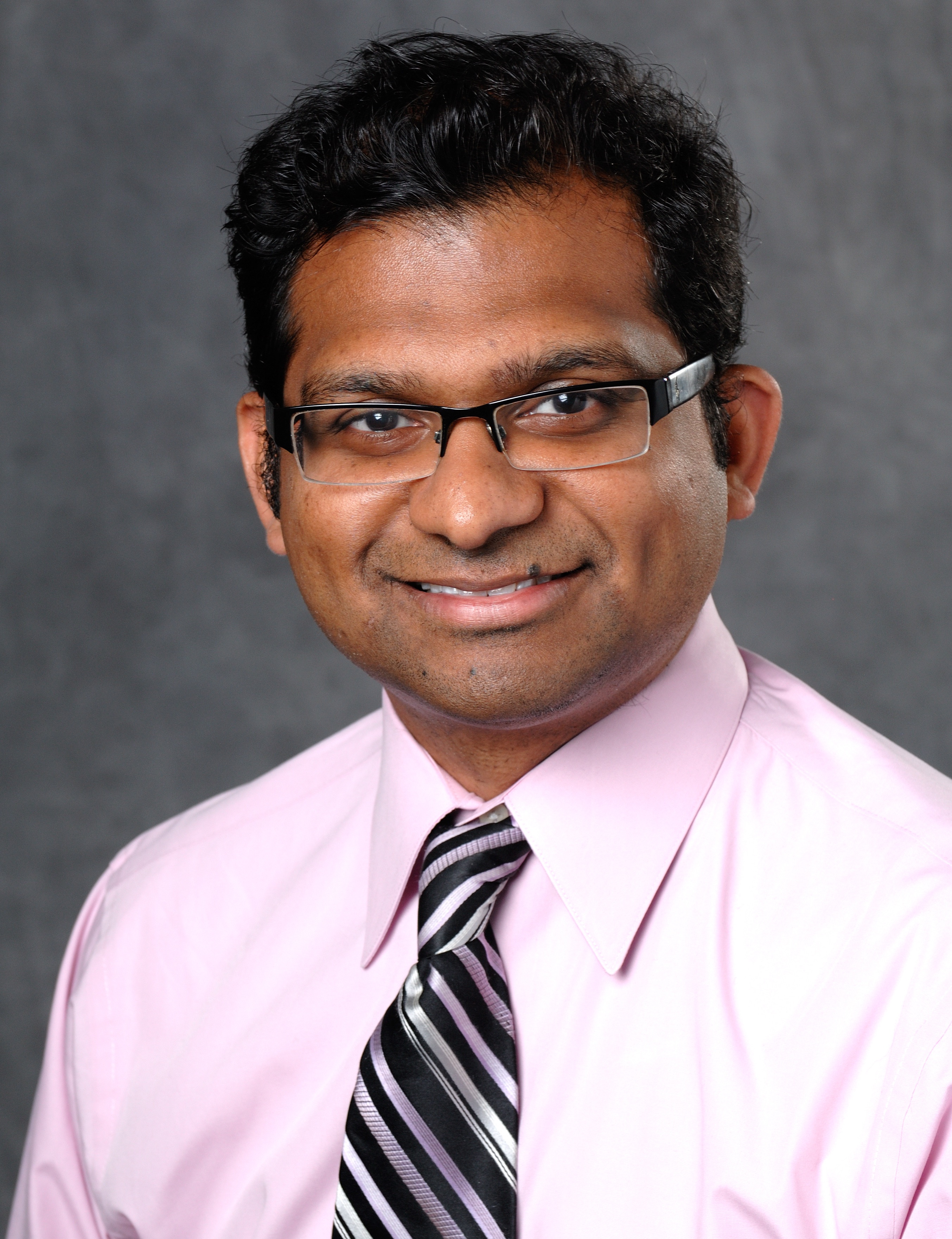}}]{Arun Ross} 
is a Professor in Michigan State University and the Director of the iPRoBe Lab. He is the co-author of the books {\em Handbook of Multibiometrics} and {\em Introduction to Biometrics}. Arun is a recipient of the IAPR JK Aggarwal Prize, IAPR Young Biometrics Investigator Award and the NSF CAREER Award.
\end{IEEEbiography}






\end{document}